\documentclass[final,3p,times]{elsarticle}

\usepackage{url}

\usepackage{adjustbox}
\usepackage[utf8]{inputenc}

\usepackage{graphicx}%
\usepackage{multirow}%
\usepackage{amsmath,amssymb,amsfonts}%
\usepackage{amsthm}%
\usepackage{mathrsfs}%
\usepackage[title]{appendix}%

\usepackage[dvipsnames]{xcolor}
\usepackage{xcolor}%
\usepackage{textcomp}%
\usepackage{manyfoot}%
\usepackage{booktabs}%
\usepackage{algorithm}%
\usepackage{algorithmicx}%
\usepackage{algpseudocode}%
\usepackage{listings}%
\usepackage{natbib}%
\usepackage{tabularx}
\usepackage{tabulary}
\usepackage{threeparttable} 
\usepackage{comment}

\usepackage{hyperref}

\usepackage{times}
\usepackage{epsfig}
\usepackage{multicol}
\usepackage{makecell}
\usepackage{enumitem}
\usepackage{array}
\usepackage{siunitx}
\usepackage{mathtools}
\usepackage{nomencl}
\usepackage{soul}
\usepackage[usestackEOL]{stackengine}
\usepackage{soul}

\usepackage{listings}
\definecolor{codegreen}{rgb}{0,0.6,0}
\definecolor{codegray}{rgb}{0.5,0.5,0.5}
\definecolor{codepurple}{rgb}{0.58,0,0.82}
\definecolor{backcolour}{rgb}{0.95,0.95,0.92}

\lstdefinestyle{mystyle}{
    backgroundcolor=\color{backcolour},   
    commentstyle=\color{codegreen},
    keywordstyle=\color{magenta},
    numberstyle=\tiny\color{codegray},
    stringstyle=\color{codepurple},
    basicstyle=\ttfamily\footnotesize,
    breakatwhitespace=false,         
    breaklines=true,                 
    captionpos=b,                    
    keepspaces=true,                 
    numbers=left,                    
    numbersep=5pt,                  
    showspaces=false,                
    showstringspaces=false,
    showtabs=false,                  
    tabsize=2
}
\newcommand\scalemath[2]{\scalebox{#1}{\mbox{\ensuremath{\displaystyle #2}}}}

\usepackage[capitalize]{cleveref}
\crefname{section}{Sec.}{Secs.}
\Crefname{section}{Section}{Sections}
\Crefname{table}{Table}{Tables}
\crefname{table}{Tab.}{Tabs.}

\journal{ArXiv}

\begin{document}

\begin{frontmatter}

\title{Role of Mixup in Topological Persistence Based Knowledge Distillation for Wearable Sensor Data}

\author[label1]{Eun~Som~Jeon}
\author[label2]{Hongjun~Choi}

\author[label3]{Matthew P. Buman}
\author[label4]{Pavan~Turaga}

\affiliation[label1]{organization={Department of Computer Science and Engineering, Seoul National University of Science and Technology},
             city={Seoul},
             postcode={01811},
             country={South Korea}}
\affiliation[label2]{organization={Lawrence Livermore National Laboratory},
             city={Livermore},
             postcode={94550},
             state={CA},
             country={USA}}
\affiliation[label3]{organization={College of Health Solutions, Arizona State University},
             city={Phoenix},
             postcode={85004},
             state={AZ},
             country={USA}}
\affiliation[label4]{organization={Geometric Media Lab, School of Arts, Media and Engineering and School of Electrical, Computer and Energy Engineering, Arizona State University},
             city={Tempe},
             postcode={85281},
             state={AZ},
             country={USA}}

\begin{abstract}
The analysis of wearable sensor data has enabled many successes in several applications. To represent the high-sampling rate time-series with sufficient detail, the use of topological data analysis (TDA) has been considered, and it is found that TDA can complement other time-series features. 
Nonetheless, due to the large time consumption and high computational resource requirements of extracting topological features through TDA, it is difficult to deploy topological knowledge in machine learning and various applications. In order to tackle this problem, knowledge distillation (KD) can be adopted, which is a technique facilitating model compression and transfer learning to generate a smaller model by transferring knowledge from a larger network. By leveraging multiple teachers in KD, both time-series and topological features can be transferred, and finally, a superior student using only time-series data is distilled.
On the other hand, mixup has been popularly used as a robust data augmentation technique to enhance model performance during training.
Mixup and KD employ similar learning strategies. In KD, the student model learns from the smoothed distribution generated by the teacher model, while mixup creates smoothed labels by blending two labels. Hence, this common smoothness serves as the connecting link that establishes a connection between these two methods.
Even though it has been widely studied to understand the interplay between mixup and KD, most of them are focused on image based analysis only, and it still remains to be understood how mixup behaves in the context of KD for incorporating multimodal data, such as both time-series and topological knowledge using wearable sensor data.
In this paper, we analyze the role of mixup in KD with time-series as well as topological persistence, employing multiple teachers. We present a comprehensive analysis of various methods in KD and mixup, supported by empirical results on wearable sensor data. We observe that applying mixup to training a student in KD improves performance.
We suggest a general set of recommendations to obtain an enhanced student.
\end{abstract}


\end{frontmatter}



\section{Introduction}\label{sec1}
Wearable sensor data analysis has enabled many application by utilizing the power of deep learning. However, there are common challenges, such as inter- and intra-person variability, sensor-level
noises, dependency on the sampling rate of the sensors, resulting in performance degradation and difficulties for deployment with machine learning. To mitigate these problems, topological data analysis (TDA) methods have been utilized on wearable sensor data analysis \cite{nawar2020topological, som2020pi, ejasilomar}, which have resulted in many robust ways to capture detailed time-series information, and can be increasingly applied to many different areas.
TDA methods allow for capturing and preserving shape-related information and have the potential to make sensor data processing pipelines more robust to different types of time-series corruptions \cite{adams2017persistence,seversky2016time,Munch2017}. Topological features can be represented in many ways \cite{Barnes2021,edelsbrunner2022computational}, a common approach is referred to as the persistence image (PI) -- which can aid in easily deploy topological persistence in machine learning owing to it 2D image-like form. Prior research has found that persistence images provide additional information that complements the raw time-series data to improve performance in time-series classification problems on wearable sensor data \cite{som2020pi, ejasilomar, 10308705cadtp}. 
Applications of topological methods also have touched upon many areas particularly in sensor data analysis \cite{rieck2020uncovering,Jiang2022,Yan2022}. 

Although TDA has shown great promise, leveraging topological features by TDA on edge-devices including wearable devices, particularly implementing them on small form factor and memory limited devices, is difficult because of large computational resources and time consumption requirements to extract the topological features \cite{adams2017persistence, Chazal2021}. Also, previous studies implement separate models in test-time simultaneously to utilize topological as well as time-series data to improve performance \cite{som2020pi}, which can increase the complexity of a model. Based on this insight, new methods to create a unified model for maximizing efficiency and integration of topological features is required.

To address these issues, knowledge distillation (KD) can be adopted as a solution, which generates a small and superior model by transferring knowledge from a large network model. Furthermore, it enables to leverage multimodal data to distill a robust single model. With KD, a teacher trained with topological features can be utilized to provide more diverse information to a student while complementing time-series features. With multiple teachers trained with the raw time-series and topological representations, a single and superior student, using the time-series data alone, can be distilled \cite{ejasilomar}. 

In KD, the temperature hyperparameter plays a key role in learning process, 
which controls the smoothness of distribution
and determines the difficulty level of the distillation process.
In this context, recently, many studies have delved into the impact of mixup augmentation in KD \cite{NEURIPS2022_57b53238, choi2023understanding, li2021smile, yang2022mixskd, xu2023computation}.
Particularly, for image analysis, Choi \emph{et al.} \cite{choi2023understanding} explored the interplay of mixup with KD and revealed that smoothness serves as the connecting link to understand the effect of mixup in KD. For more details, in KD, the student learns from the smoothed distribution provided by the teacher model, and this distribution is further smoothed by increasing the temperature value. Similarly, mixup generates new smooth labels by blending two given inputs and ground truth labels, which are then further smoothed by strongly interpolated samples (e.g., a high alpha value in the beta distribution).
Thus, their behave as a connecting link for promoting smoothness in learning process, which can generate synergetic effects to distill a robust lightweight model \cite{choi2023understanding, yang2022mixskd}.

There are different augmentation methods such as regularization effect \cite{bishop1995training}, model invariance \cite{chen2020group}, and feature learning \cite{shen2022data}. 
However, these techniques are more focus on alleviating noises or data point issues in rotation, which are different from mixup \cite{allenzhu2023towards} blending multiple samples. Further, even if other augmentations (e.g. cutmix \cite{yun2019cutmix} and adversarial training \cite{allen2022feature}) are effective, mixup offers different benefits in much lower computational overhead and provides solid foundations, particularly in the context of knowledge distillation \cite{zhao2021data, zou2023benefits, beyer2022knowledge}.


Even though the interplay between two techniques, mixup and KD, is significantly crucial in performance improvement, the majority of previous studies have primarily concentrated on image-based analysis. To the best of our knowledge, the impact of mixup and KD in the context of both time-series and topological representations on wearable sensor data remains unexplored.  
Furthermore, the behavior of mixup for multiple teachers and different strategies in KD have not been investigated.



In this paper, we study the behavior of mixup in KD with multimodalities using both time-series and topological representations for wearable sensor data analysis. We implement different KD approaches for utilizing time-series as well as topological persistence to train a student. 
We investigate whether the mixup method can enhance the performance of topological persistence-based KD using various teachers. Additionally, we compare the performance of using mixup in KD to determine if leveraging both representations yields more benefits than relying solely on time-series data.


The contributions of this paper are summarized below:

\begin{itemize}
    \item We analyze the interplay between mixup and KD for wearable sensor data, and compare different strategies in KD with single-teacher and multiple-teacher based distillation, leveraging time-series as well as topological persistence.
    \item We study the effects of mixup on training both teacher and student models. We aim to identify which training strategy for utilizing mixup in KD provides the most benefit in the activity classification task and explore whether the effects of mixup are comparable to those of other time domain augmentation methods in KD.
    \item Through the analysis of multiple strategies for employing mixup with multiple teachers, we propose improved learning approaches by regulating smoothness through temperature and the number of mixup pairs.
\end{itemize}

The rest of the paper is organized as follows. In section \ref{sec:background}, we describe mixup and KD techniques with persistence image. In section \ref{sec:strategies}, we explain strategies to leverage topological persistence with mixup in KD.
In section \ref{sec:exp}, we present our experimental results and analysis. In section \ref{sec:conclu}, we discuss our findings and conclusions.

\section{Background}
\label{sec:background}

\subsection{Mixup Augmentation}
Mixup augmentation \cite{zhang2018mixup} is used commonly in deep-learning techniques to alleviate issues of memorization and sensitivity to adversarial examples. Two examples drawn at random from training data are mixed by linear interpolation \cite{zhang2018mixup}. Let the training data be $\mathcal{D}=\{(x_{1},y_{1}),...,(x_{n},y_{n})\}$, where $n$ is the number of samples. Input data is $x \in \mathcal{X} \subseteq \mathbb{R}^{d}$ and its corresponding label is $y \in \mathcal{Y}=\{1,2,...,K\}$. 
The sampling process for mixup can be written as follows:
\begin{equation}
\begin{split}
&\tilde{x}_{ij}(\lambda)=\lambda x_{i} + (1-\lambda)x_{j}, \\
&\tilde{y}_{ij}(\lambda)=\lambda y_{i} + (1-\lambda)y_{j},
\end{split}
\end{equation}
\noindent where $\lambda\in [0,1]$ follows the distribution $P_{\lambda}$ where $\lambda \sim \text{Beta($\alpha$, $\alpha$)}$. $\lambda$ is to specify the extent of mixing. The hyper-parameter $\alpha$ controls the strength of interpolation between feature-target pairs. $\alpha$ generates strongly interpolated samples. To train a function $f$, the following mixup loss function is minimized:
\begin{equation}\label{eq:mixup_eq}
\mathcal{L}_{mix}(f)= \frac{1}{n^{2}}\sum_{i=1}^{n}\sum_{j=1}^{n}\mathbb{E}_{\lambda\sim P_{\lambda}}[\mathcal{L}_{CE}(f(\tilde{x}_{ij}(\lambda)), \tilde{y}_{ij}(\lambda))],
\end{equation}
\noindent where $\mathcal{L}_{CE}$ is a standard cross-entropy loss function.

Many different variants of mixup have been studied \cite{verma2019manifold, yun2019cutmix, kim2021co}. Intrinsically, these methods have similarities in that they mix the input data (e.g. images) and labels proportionally to extend the training distribution. The benefits of mixup with time-series data were explored in previous studies \cite{darlow2023tsmix, aggarwal2023embarrassingly, Zhou2023}. In this study, we use the conventional mixup to explore the effects on knowledge distillation \cite{zhang2018mixup} for time-series data.

\subsection{Persistence Image}
\begin{figure}[htb!]
\includegraphics[scale=0.34] {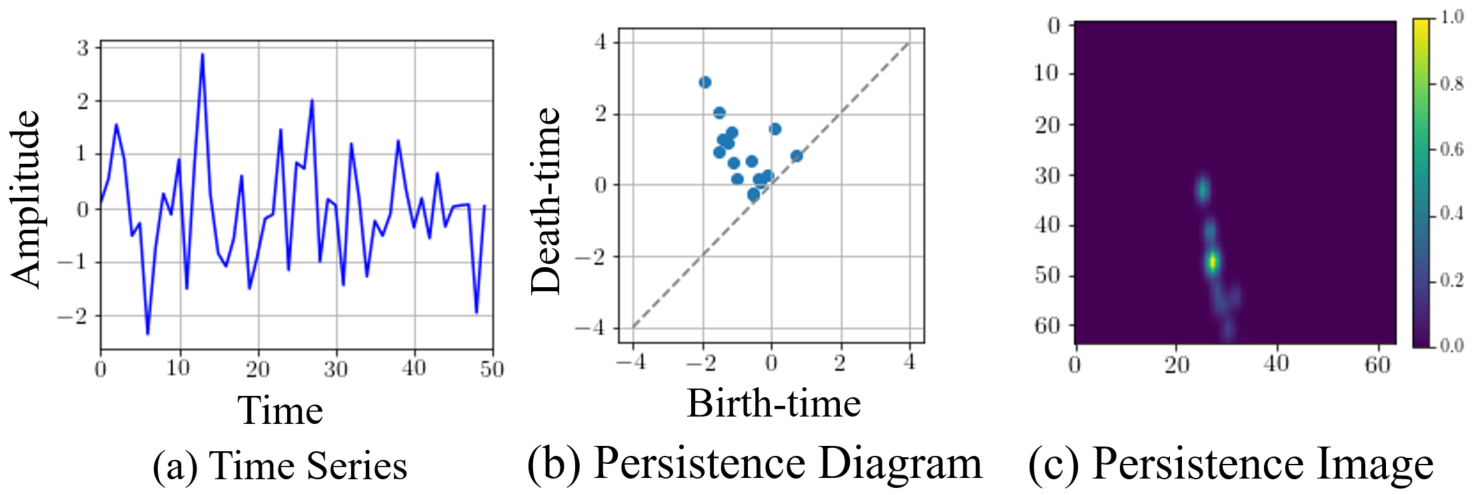} 
\centering
\caption{time-series data and its corresponding PD and PI. Higher persistence in PD is represented with brighter color in PI.}
\label{figure:ts_pd_pi}
\end{figure}

TDA has been applied in various fields \cite{adams2017persistence, WANG2021109324, gholizadeh2018short, zeng2021topological}, which can characterize the shape of raw data. One important tool in TDA is persistent homology, which provides a multiscale description with topological features. When applied to point clouds, these features are often described as cavities characterized by points, triangles, and edges by filtration  \cite{stolz2023outlier, edelsbrunner2022computational}. The extension to time-series data is via sub level-set filtrations, where level-sets are tracked. The birth and death times of topological features can be represented as a multiset of points in a persistence diagram (PD). Since the number and locations of the points in PDs vary depending on the underlying data, it is difficult to use them directly in machine learning pipelines. To project the features on the stable vector representation, a persistence image can be used, mapping the scatter points based on their persistence value (life time) \cite{adams2017persistence}. Firstly, PD is mapped to an integrable function $\rho: \mathbb{R} \rightarrow \mathbb{R}^2$, called a persistence surface (PS), which is defined as a weighted sum of Gaussian functions. A PI can be created by integrating PS on a grid box that is defined by discretization. The values of PI represent the persistence points of the PD. The example of PD and PI are shown in Fig. \ref{figure:ts_pd_pi}. Even though TDA can provide additional information to the raw time-series to improve performance, it is challenging to run the method on a resource constrained devices, because extracting PIs by TDA requires a large amount of time and memory. To solve this problem, in this paper, we adopt knowledge distillation that distills a single student utilizing the raw time-series data alone as an input.

\subsection{Knowledge Distillation}
Knowledge distillation trains a smaller (student) model from a larger (teacher) model \cite{bucilua2006model, hinton2015distilling}.
The student model is trained by minimizing the difference between its outputs and soft labels, called relaxed knowledge, from a teacher, which improves performance beyond using hard labels (labeled data) alone.
The loss function of standard knowledge distillation \cite{hinton2015distilling} is:
\begin{equation}
\label{eq: stdkd_loss}
\mathcal{L} = (1-\tau)\mathcal{L}_{CE}(\sigma(t_s), y_g) + \tau \mathcal{L}_{KD}(f_{T}, f_{S}),\\
\end{equation}
where $t_s$ is logits of a student model $f_{S}$, $f_{T}$ is a teacher model, $y_g$ is a ground truth label, $\sigma(\cdot)$ is a softmax function, $\mathcal{L}_{KD}(\cdot)$ is KD loss function, and $\tau$ is hyper-parameter; $0 < \tau < 1$.
The difference between the outputs of the student and the teacher is mitigated by employing Kullback-Leibler divergence loss function, which is described as follows:
\begin{equation}\label{kld_loss}
    \mathcal{L}_{KD}(f_{T}, f_{S}) = \frac{\mathcal{T}^{2}}{n}\sum_{i=1}^{n}KL(\sigma(\frac{f_{T}(x_{i})}{\mathcal{T}}), \sigma(\frac{f_{S}(x_{i})}{\mathcal{T}})),
\end{equation}
where $KL$($\cdot$) measures Kullback-Leibler divergence loss, $\mathcal{T}$ is a hyper-parameter, temperature, to smooth the outputs. To obtain the best performance, in this paper, we utilize a teacher trained by early stopping the training process in KD \cite{cho2019efficacy}.

Not only logits, but also features from intermediate layers can be utilized to knowledge transfer, which is called feature-based distillation  \cite{gou2021knowledge}. 
Attention transfer (AT) has been widely used, which uses attention maps extracted by a sum of squared attention mapping function \cite{zagoruyko2016paying}.
Tung \emph{et al.} \cite{tung2019similarity} extracts similarities within a mini-batch of samples from a teacher and a student, where those maps have to be matched in distillation process.
Even though various techniques have been utilized to improve the performance, they typically address single-modal issues with a single teacher.

Multiple teachers can be utilized to provide more and diverse knowledge to a single student \cite{ejasilomar, gou2021knowledge, liu2020adaptive, zhang2022confidence}. Using a uni-modal data with different teachers, a student can establish its own knowledge by integrating diverse knowledge from the teachers \cite{you2017learning}.
However, in some cases, data samples or labels used for training a teacher cannot be leveraged to train or test a student \cite{gou2021knowledge}. 
Jeon \emph{et al.} \cite{ejasilomar} utilize multiple teachers to train a single student by transferring features from both the persistence image and the raw time-series data. Even though two teachers have different architectural designs and use different types of inputs, their logit information can be transferred with KD loss that can be written as:
\begin{equation}
\begin{split}
    \mathcal{L}_{KDm}(f_{T_1}, f_{T_2}, f_{S}) &
    = \eta \mathcal{L}_{KD}(f_{T_1}, f_{S}) \\
    &+ (1-\eta) \mathcal{L}_{KD}(f_{T_2}, f_{S}),
\end{split}
\end{equation}
where $\eta$ is a hyper-parameter to control the effects from different teachers, and $f_{T_1}$ and $f_{T_2}$ are teacher models trained with time-series data and PIs, respectively. Then, the total loss function can be written as:
\begin{equation}
    \mathcal{L}_{m} = (1-\tau)\mathcal{L}_{CE}(\sigma(t_s), y_g) + \tau \mathcal{L}_{KDm}(f_{T_1}, f_{T_2}, f_{S}).
\end{equation}

For further improvement in KD, mixup augmentation methods have been widely studied. Specifically, mixup and KD share a common thread in serving smoothness during the training process. 
To accommodate synergetic effects, 
the interest in the interplay between mixup and KD grows, which has been analyzed in many studies \cite{NEURIPS2022_57b53238, choi2023understanding, li2021smile, yang2022mixskd, xu2023computation}. However, most of the studies were conducted with image data only. It is still required to be explored with time-series and multimodalities using different representations. Based on these insights, we investigate the effects of mixup in KD for time-series on wearable sensor data by utilizing a single or multiple teachers. Also, we present compatible or incompatible views through an empirical analysis.

\section{Analysis Strategies for Mixup in KD}\label{sec:strategies}

To analyze the effect of mixup in persistence based KD, we utilize different approaches that are explained in this section.

\subsection{Leveraging Topological Persistence}

\begin{figure}[htb!]
\includegraphics[scale=0.62] {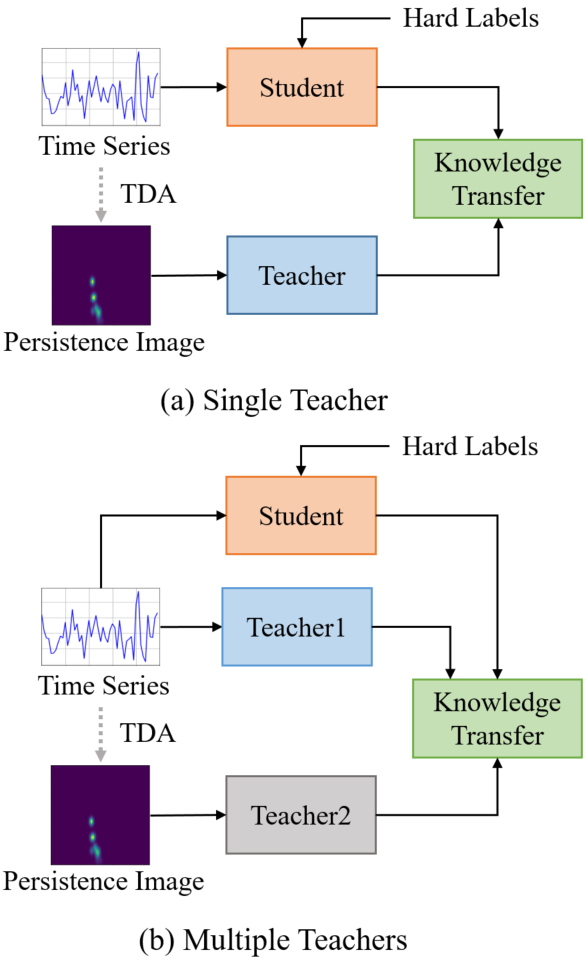} 
\centering
\caption{Strategies to leverage topological persistence in KD. (a) utilizes a single teacher trained with PIs. (b) uses different teachers trained with PIs and the raw time-series data, respectively.}
\label{figure:tda_kd}
\end{figure}

\subsubsection{Leveraging A Single Teacher}
With the process of standard knowledge distillation, a single teacher trained with PIs can be used to transfer knowledge to a student, as illustrated in Fig. \ref{figure:tda_kd}(a). PIs are generated by TDA from the raw time-series data. PIs are 2D images, so the teacher model consists of a 2D kernel of CNNs. To train a student with time-series (1D) data, 1D CNNs can be used. Logit of the teacher and student is leveraged to transfer knowledge.

\subsubsection{Leveraging Multiple Teachers}
Multiple teachers can be used to train a single student. For instance, two teachers, trained with time-series and PIs, can transfer knowledge simultaneously, as described in Fig. \ref{figure:tda_kd}(b). The student utilizes time-series alone as an input. In this way, the student can obtain benefits from both of these different features, but it still requires only time-series implementation at test time. Since two teachers are trained with different modalities and have different architectural designs, it is difficult to create a unified model and knowledge gap making performance degradation can be produced \cite{gou2021knowledge}. To mitigate this issue, we adopt an annealing strategy that trains a student by initializing weight values from a model learned from scratch \cite{ejasilomar}.

\subsection{Mixup Strategy in KD}

\begin{figure*}[htb!]
\includegraphics[scale=0.58] {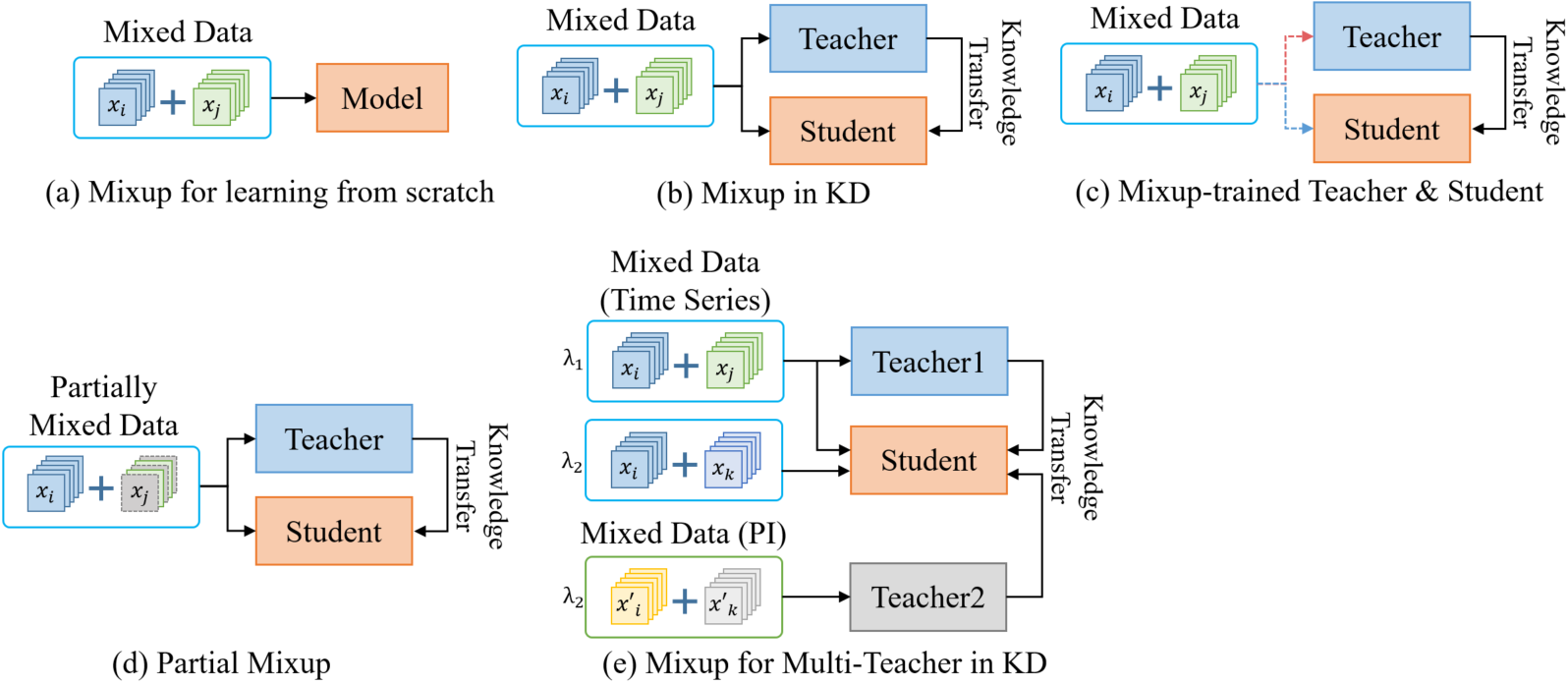} 
\centering
\caption{Approaches for incorporating mixup in KD.}
\label{figure:mixup_strategy}
\end{figure*}

We set different strategies to utilize mixup in KD, as described in Fig. \ref{figure:mixup_strategy}. Details are explained as follows.

\begin{itemize}
\item \textbf{Mixup for learning from scratch:} To investigate the effects of mixup on time-series, we compare mixup- and non-mixup trained models.

\item \textbf{Mixup in KD:} To explore the connecting link between mixup and KD, we train a student model with mixup and different temperatures, using various methods in KD.

\item \textbf{Mixup-trained teacher and student:} We apply mixup not only to a student but also to teachers to figure out the effects of the augmentation method in KD. With different combinations of applying mixup, we investigate which strategy is effective in KD.

\item \textbf{Distillation with different temperature and partial mixup:} To analyze the effects of smoothness from temperature on mixup in KD, a student is trained with the augmentation method and different temperature parameters. In this way, we figure out how much temperature impacts the performance of mixup in KD. Also, to analyze the smoothness of mixup, we utilize partial mixup (PMU) that uses only a few mixup pairs in a batch, as addressed in the previous study \cite{choi2023understanding}. The method uses small amounts of mixup pairs to control the strength of smoothness, which alleviates excessive smoothness.

\item \textbf{Mixup for different teachers:} Two teachers generate different knowledge and effects for a student in distillation. To explore the effects of mixup for different modalities, we apply different hyper-parameters to teachers. The training objective for the student in KD with multiple teachers and different mixup hyper-parameters is as follows:
\begin{equation}\label{eq:final_loss_eq2}
\scalemath{0.87}{
\begin{split}
&\min\mathbb{E}_{(x,y)\sim\mathcal{D}}\left[ \right. \\
&\left. \mathbb{E}_{\lambda_1\sim P_{\lambda_1}}[\eta\{(1-\tau)\mathcal{L}_{mix}(f_S) + \tau\mathcal{L}_{KD}({f}_{T_1}, f_S)\}] + \right. \\
&\left. \mathbb{E}_{\lambda_2\sim P_{\lambda_2}}[(1-\eta)\{(1-\tau)\mathcal{L}_{mix}(f_S) + \tau\mathcal{L}_{KD}({f}_{T_2}, f_S)\}] \right],
\end{split}
}
\end{equation}
where $\lambda_1$ and $\lambda_2$ are to specify the extent of mixing, whose $\alpha$ parameters are different.
\end{itemize}

In Table \ref{table:flop}, we provide the floating point operations per second (FLOPs) with networks and processing time for an epoch with batch size of 64 in training process for strategies in Fig. \ref{figure:mixup_strategy}.
The processing time is measured on a desktop with a 3.50 GHz CPU (Intel® Xeon(R) CPU E5-1650 v3), 48 GB memory, and NVIDIA TITAN Xp (3840 NVIDIA® CUDA® cores and 12 GB memory) graphic card.
As explained in the table, Strategy (e) takes the longest time and larger complexity compared to other strategies. 
Through the training, all of strategies distill the same sized single student even though each strategy is different. In test-time, a single student model is implemented alone, which corresponds to the Student.

More details of settings and experimental results for each strategy are explained in section \ref{sec:exp}.

\begin{table}[htb!]
\centering
\caption{Details of efficiency for different training strategies with mixup and KD, which are explained in Fig. \ref{figure:mixup_strategy}. Teachers are WRN16-3 and Student is WRN16-1.}

\begin{center}
\begin{tabular}{c |c  c | c |c }

\hline
\centering

\multirow{2}{*}{Strategy} & \multicolumn{3}{c|}{GFLOPs}  & Processing \\ \cline{2-4}
& \multicolumn{2}{c|}{Teacher} & Student & Time (sec) \\ \hline
(a) & \multicolumn{2}{c|}{--} & \multirow{4}{*}{0.71} & 4.54 \\ \cline{2-3} \cline{5-5}
(b) & \multicolumn{2}{c|}{\multirow{3}{*}{6.02}} & & \multirow{2}{*}{8.19} \\ 
(c) & & & & \\ \cline{5-5}
(d) & & & & 8.50\\ \hline\hline
\multirow{2}{*}{Strategy} & \multicolumn{3}{c|}{GFLOPs}  & Processing \\ \cline{2-4}
 & Teacher1 & Teacher2 & Student & Time (sec) \\ \hline
(e) &  6.02 & 57.55 & 0.71 & 22.48 \\
\hline

\end{tabular}
\end{center}
\label{table:flop}
\end{table}

\section{Experiments} \label{sec:exp}
In this section, we describe datasets and implementation details. We utilize various strategies of KD and mixup to investigate the effects on wearable sensor data analysis. We analyze optimized solutions and describe ablations.


\subsection{Dataset Description and Implementation Details}
\subsubsection{Dataset Description}
We analyze the strategies with wearable sensor data on GENEActiv and PAMAP2 datasets. These datasets consist with diverse window size and number of channels obtained from multiple sensors on different activities. Thus, experiments on these datasets aid in showing various evaluations under different conditions, which helps to explain generalizability and applicability of methods.

\textbf{GENEActiv.} GENEActiv dataset \cite{wang2016statistical} was collected by GENEActiv sensor, using waterproof, a light-weight and writ-worn tri-axial accelerometer. The sampling frequency was 100 Hz. By referring to the previous study \cite{jeon2022kd, ejasilomar}, we select 14 daily activities for analysis, such as walking, standing, and sitting. Each class has over 9 hundred samples with 500 time steps of window size, corresponding to 5 seconds with full-non-overlapping sliding windows. The number of subjects for training and testing is 130 and 43, respectively, and the number of samples is around 16k and 6k, respectively.

\textbf{PAMAP2.} PAMAP2 dataset \cite{reiss2012introducing} was recorded from heart rate, temperature, accelerometers, gyroscopes, and magnetometers, which include 3 Colibri wireless inertial measurement units (IMU). The sampling frequency was 100 Hz for 9 subjects. The recordings are downsampled to 33.3Hz by referring to the previous study \cite{jordao2018human, jeon2022kd}. A window size for a sample is 100 time steps or 3 seconds with 22 time steps for segmenting the sequences, which allows semi-non-overlapping sliding windows with 78\% overlapping \cite{reiss2012introducing}. We use 12 daily activities including lying, sitting, walking, etc. For evaluation in experiments, we use leave-one-subject-out combinations.

\subsubsection{Implementation Details}
We use the Scikit-TDA python library \cite{scikittda2019} and the Ripser package to produce PDs and extract PIs \cite{som2020pi}. For GENEActiv, the standard deviation for the Gaussian kernel is set to 0.25 and the birth-time range of PI is $[$-10, 10$]$, respectively, as do the same in the previous studies \cite{ejasilomar, som2020pi}. For PAMAP2, the parameter for Gaussian kernel is 0.015 and the range for PI is $[$-1, 1$]$, respectively. Each PI is generated from each channel and the values are normalized by its maximum intensity value. The size of PI is set to 64$\times$64. For training models, we set the total number of epochs as 200, SGD with momentum of 0.9, a weight decay of $1\times10^{-4}$, and batch size for 64. To train a model with time-series data (1D data), 1D convolutional layers are utilized. The initial learning rate is 0.05 that decreases by 0.2 at 10 epochs and drops by 0.1 every [${e \over 3}$] where $e$ is the total number of epochs. A model using image representation for PIs consists of 2D convolutional layers. The initial learning rate is 0.1 that drops by 0.5 at 10 epochs and by 0.2 at every 40 epochs. We measure the performance with WideResNet (WRN) \cite{zagoruyko2016wide} that is popularly utlized in the validation of KD \cite{cho2019efficacy, jeon2022kd, ejasilomar}. For default settings, we set $\tau$, $\eta$, and $\mathcal{T}$ as 0.7, 0.7, and 4 for GENEActiv, and 0.99, 0.3, and 4 for PAMAP2, referring to the previous study \cite{jeon2022kd, ejasilomar} and to consider best performance. We run 3 times and report the averaged accuracy and standard deviation. As a baseline, we implement standard KD \cite{hinton2015distilling}, attention transfer (AT) \cite{komodakis2017paying}, and similarity-preserving knowledge distillation (SP) \cite{tung2019similarity}, which utilize logit as well as feature from intermediate layers for distillation. Parameters for AT and SP are set as 1500 and 1000 for GENEActiv, and 3500 and 700 for PAMAP2, respectively. A simple knowledge distillation (SimKD) \cite{chen2022knowledge} and DIST \cite{huang2022knowledge} leveraging intra- and inter-class relations for knowledge transfer are also used as baselines.
Also, multi-teacher based approaches such as AVER \cite{you2017learning}, EBKD \cite{kwon2020adaptive}, and CA-MKD \cite{zhang2022confidence}, Base \cite{ejasilomar} are used for baselines. Since two teachers are incorporated with different dimensional layers, only logits are used for distillation of baselines. When mixup is applied, $\alpha$ is 0.1 for both datasets.


\begin{table}[htb!]
\centering
\renewcommand{\tabcolsep}{1.0mm} 
\caption{Accuracy ($\%$) with various knowledge distillation methods on GENEActiv.}

\begin{center}
\begin{tabular}{c c c|c |c |c |c }

\hline
\centering

\multirow{2}{*}{\shortstack{Teacher1 \\ (1D CNNs)}}& \multirow{2}{*}{\shortstack{Teacher2 \\ (2D CNNs)}} & \multirow{2}{*}{\shortstack{Student \\ (1D CNNs)}} & TS & PI & \multicolumn{2}{c}{TS+PI} \\
\cline{4-7}
 & & & KD & KD & Base & Ann. \\
\hline

WRN16-1 & WRN16-1 &  \multirow{12}{*}{\shortstack{WRN16-1 \\ (0.06M \\ 67.66)}} & \multirow{3}{*}{\shortstack{69.71 \\ {\scriptsize$\pm$0.38}}} & \multirow{3}{*}{\shortstack{67.83 \\ {\scriptsize$\pm$0.17}}} & \multirow{3}{*}{\shortstack{69.09 \\ {\scriptsize$\pm$0.37}}} & \multirow{3}{*}{\shortstack{\textbf{70.15} \\ {\scriptsize$\pm$0.03}}} \\
(0.06M, & (0.2M, &   & & & & \\
67.66) & 58.64) &  & & & &  \\

WRN16-3 & WRN16-3 & & \multirow{3}{*}{\shortstack{69.50 \\ {\scriptsize$\pm$0.10}}} & \multirow{3}{*}{\shortstack{68.79 \\ {\scriptsize$\pm$0.73}}} & \multirow{3}{*}{\shortstack{69.24 \\ {\scriptsize$\pm$0.62}}} & \multirow{3}{*}{\shortstack{\textbf{70.71} \\ {\scriptsize$\pm$0.12}}} \\
(0.5M, & (1.6M, &   & & & & \\
68.89) & 59.80) &  & & & &  \\

WRN28-1 & WRN28-1 &  & \multirow{3}{*}{\shortstack{68.32 \\ {\scriptsize$\pm$0.63}}} & \multirow{3}{*}{\shortstack{68.51 \\ {\scriptsize$\pm$0.01}}} & \multirow{3}{*}{\shortstack{69.55 \\ {\scriptsize$\pm$0.41}}} & \multirow{3}{*}{\shortstack{\textbf{70.44} \\ {\scriptsize$\pm$0.10}}} \\
(0.1M, & (0.4M, &  & & & & \\
68.63) & 59.45) & & & & &  \\

WRN28-3 & WRN28-3 &  & \multirow{3}{*}{\shortstack{68.01 \\ {\scriptsize$\pm$0.69}}} & \multirow{3}{*}{\shortstack{68.46 \\ {\scriptsize$\pm$0.28}}} & \multirow{3}{*}{\shortstack{69.42 \\ {\scriptsize$\pm$0.58}}} & \multirow{3}{*}{\shortstack{\textbf{69.97} \\ {\scriptsize$\pm$0.06}}} \\
(1.1M, & (3.3M, &  & & & & \\
69.23) & 59.69) & & & & &  \\

\hline

\end{tabular}
\end{center}
\label{table:baseGENE}
\end{table}



\begin{table}[htb!]
\caption{Accuracy ($\%$) for related methods on GENEActiv with 7 classes. For KD, teachers are WRN16-3 and students are WRN16-1.}\label{table:GENE_7cls}
\centering
\begin{tabular}{c | p{7em} |c c}
\hline
\multicolumn{2}{c|}{\multirow{2}{*}{Method}} & \multicolumn{2}{c}{Window length} \\
   \multicolumn{2}{c|}{}  & 1000 & 500 \\ \hline

    \multirow{8}{*}{\rotatebox[origin=c]{90}{\scriptsize{TS}} }
    & Student & 89.29{\scriptsize$\pm$0.32} & 86.83{\scriptsize$\pm$0.15} \\  \cline{2-4}
    & SVM \cite{cortes1995support} & 86.29 & 85.86  \\
    & Choi \textit{et al.} \cite{choi2018temporal} & 89.43 & 87.86 \\  \cline{2-4}
    & KD & 89.88{\scriptsize$\pm$0.07} & 88.16{\scriptsize$\pm$0.15} \\
    & AT & 90.32{\scriptsize$\pm$0.09} & 87.60{\scriptsize$\pm$0.22}  \\
    & SP & 88.47{\scriptsize$\pm$0.19} & 87.69{\scriptsize$\pm$0.18}  \\
    & DIST & 90.20{\scriptsize$\pm$0.39} & 87.05{\scriptsize$\pm$0.31} \\
    & SimKD & 90.47{\scriptsize$\pm$0.32} & 88.16{\scriptsize$\pm$0.37} \\
    \hline
    \multirow{4}{*}{\rotatebox[origin=c]{90}{\scriptsize{TS+PI}} }
    & AVER & 90.06{\scriptsize$\pm$0.33} & 87.05{\scriptsize$\pm$0.37}  \\ 
    & EBKD & 89.82{\scriptsize$\pm$0.14} & 87.66{\scriptsize$\pm$0.28}  \\ 
    & CA-MKD & 90.13{\scriptsize$\pm$0.34} & 88.04{\scriptsize$\pm$0.26}  \\ 
    & Ann. & \textbf{90.71}{\scriptsize$\pm$0.15} & \textbf{88.26}{\scriptsize$\pm$0.24}  \\  \hline
\end{tabular}
\end{table}



\begin{table}[htb!]
\centering
\renewcommand{\tabcolsep}{1.0mm} 
\caption{Accuracy ($\%$) with various knowledge distillation methods on PAMAP2.}

\begin{center}
\begin{tabular}{c c c|c |c |c |c }

\hline
\centering

\multirow{2}{*}{\shortstack{Teacher1 \\ (1D CNNs)}}& \multirow{2}{*}{\shortstack{Teacher2 \\ (2D CNNs)}} & \multirow{2}{*}{\shortstack{Student \\ (1D CNNs)}} & TS & PI & \multicolumn{2}{c}{TS+PI} \\
\cline{4-7}
 & & & KD & KD & Base & Ann. \\
\hline

WRN16-1 & WRN16-1 &  \multirow{12}{*}{\shortstack{WRN16-1 \\ (0.06M, \\ 82.99)}} & \multirow{3}{*}{\shortstack{85.96 \\ {\scriptsize$\pm$2.19}}} & \multirow{3}{*}{\shortstack{85.04 \\ {\scriptsize$\pm$2.58}}} & \multirow{3}{*}{\shortstack{85.91 \\ {\scriptsize$\pm$2.32}}} & \multirow{3}{*}{\shortstack{\textbf{86.09} \\ {\scriptsize$\pm$2.33}}} \\
(0.06M, & (0.2M, &   & & & & \\
85.27) & 86.93) &  & & & &  \\

WRN16-3 & WRN16-3 &  & \multirow{3}{*}{\shortstack{86.50 \\ {\scriptsize$\pm$2.21}}} & \multirow{3}{*}{\shortstack{86.68 \\ {\scriptsize$\pm$2.19}}} & \multirow{3}{*}{\shortstack{86.18 \\ {\scriptsize$\pm$2.37}}} & \multirow{3}{*}{\shortstack{\textbf{87.12} \\ {\scriptsize$\pm$2.26}}} \\
(0.5M, & (1.6M, &   & & & & \\
85.80) & 87.23) &  & & & &  \\

WRN28-1 & WRN28-1 &  & \multirow{3}{*}{\shortstack{84.92 \\ {\scriptsize$\pm$2.45}}} & \multirow{3}{*}{\shortstack{85.08 \\ {\scriptsize$\pm$2.44}}} & \multirow{3}{*}{\shortstack{85.54 \\ {\scriptsize$\pm$2.26}}} & \multirow{3}{*}{\shortstack{\textbf{85.89} \\ {\scriptsize$\pm$2.26}}} \\
(0.1M, & (0.4M, &  & & & & \\
84.81) & 87.45) & & & & &  \\

WRN28-3 & WRN28-3 &  & \multirow{3}{*}{\shortstack{86.26 \\ {\scriptsize$\pm$2.40}}} & \multirow{3}{*}{\shortstack{85.39 \\ {\scriptsize$\pm$2.35}}} & \multirow{3}{*}{\shortstack{86.04 \\ {\scriptsize$\pm$2.34}}} & \multirow{3}{*}{\shortstack{\textbf{86.33} \\ {\scriptsize$\pm$2.30}}} \\
(1.1M, & (3.3M, &  & & & & \\
84.46) & 87.88) & & & & &  \\

\hline

\end{tabular}
\end{center}
\label{table:basePAMAP2}
\end{table}


\begin{table}[htb!]
\caption{Accuracy ($\%$) for related methods on PAMAP2. For KD, teachers are WRN16-3 and students are WRN16-1.}
\label{table:PAMAP2_comparison}
\centering
\begin{tabular}{c |p{8em} | c } 
\hline
\multicolumn{2}{c|}{Method} & Accuracy \\ \hline

\multirow{9}{*}{\rotatebox[origin=c]{90}{\scriptsize{TS}} }

& Student & 82.81{\scriptsize$\pm$2.51}  \\ \cline{2-3}

& Chen and Xue \cite{chen2015deep}  &  83.06  \\
& Ha \textit{et al.}\cite{ha2015multi}  &  73.79  \\
& Ha and Choi \cite{ha2016convolutional}  &  74.21  \\
& Catal \textit{et al.} \cite{catal2015use}  &  85.25  \\
& Kim \textit{et al.}\cite{kim2012analysis}  &  81.57  \\ \cline{2-3}

& KD & 86.38{\scriptsize$\pm$2.25} \\
& AT & 84.44{\scriptsize$\pm$2.22}  \\
& SP & 84.89{\scriptsize$\pm$2.10}  \\ \hline

\multirow{5}{*}{\rotatebox[origin=c]{90}{\scriptsize{TS+PI}} }
& AVER & 86.00{\scriptsize$\pm$2.45}  \\
& EBKD & 85.62{\scriptsize$\pm$2.37}  \\
& CA-MKD & 85.02{\scriptsize$\pm$2.64}  \\
& Base & 86.18{\scriptsize$\pm$2.37} \\
& Ann. & \textbf{87.12}{\scriptsize$\pm$2.26}  \\  \hline

\end{tabular}
\end{table}

\subsection{Preliminary: Effects of Topological Persistence in KD}
In this section, as preliminaries, we conduct experiments with a single and multiple teacher based distillation methods.
For multiple teacher based methods, we train models with time-series as well as PIs by leveraging topological persistence. Teachers and students are trained with the various KD strategies explained in the previous section. Note, ``TS'' and ``Ann.'' denote using time-series data to train a student model and using two teachers in KD and an annealing strategy \cite{ejasilomar}, respectively. Teacher1 and Teacher2 are teachers trained with time-series and persistence images, respectively.

As described in Table \ref{table:baseGENE}, for GENEActiv, Ann. using multiple teachers shows the best in all cases. Among different combinations, WRN16-3 teachers distill a superior student. To compare with previous studies, we tested a combination of teachers (WRN16-3) and students (WRN16-1) on GENEActiv utilizing different window length for 7 classes, where the combination showed the best in past studies \cite{cho2019efficacy, jeon2022kd, ejasilomar}. As shown in Table \ref{table:GENE_7cls}, Ann. outperforms previous methods. Also, as summarized in Table \ref{table:basePAMAP2} and \ref{table:PAMAP2_comparison}, for PAMAP2, Ann. outperforms methods using a single teacher and previous methods. WRN16-3 teachers for Ann. produce best performance. This represent that considering coherent characteristics of a student is important to improve performance. Specifically, training a student from weights of learning from scratch helps to alleviate the knowledge gap that makes it difficult to transfer knowledge to a student from multiple teachers.
These results show that topological features implement time-series to improve the performance.

\begin{table}[htb!]
\centering
\caption{Accuracy ($\%$) for different structure of teachers on GENEActiv.}

\begin{center}
\begin{tabular}{c |c c | c c }

\hline
\centering

\multirow{2}{*}{Method} & \multicolumn{4}{c}{Architecture Difference} \\ \cline{2-5}
 & \multicolumn{2}{c|}{Depth} &
\multicolumn{2}{c}{Width} \\
\hline

 & WRN & WRN & WRN & WRN\\
Teacher1 & 16-1 & 28-1 &
28-1 & 28-3 \\
(1D CNNs) & (0.06M, & (0.2M, &
(0.1M, & (1.1M,\\
& 67.66) & 68.63) &
68.63) & 69.23) \\ \hline

 & WRN & WRN & WRN & WRN \\
Teacher2 & 28-1 & 16-1 &
28-3 & 28-1 \\
(2D CNNs) & (0.1M, & (0.2M, &
(3.3M, & (0.4M, \\
& 59.45) & 58.64) &
59.69) & 59.45) \\

\hline

Student & \multicolumn{4}{c}{WRN16-1}  \\ 
(1D CNNs) & \multicolumn{4}{c}{(0.06M, 67.66{\scriptsize$\pm$0.45})} \\ \hline

\multirow{2}{*}{Base} & 68.71 & 67.89 &
68.26 & 69.09 \\
 & {\scriptsize$\pm$0.36} & {\scriptsize$\pm$0.27} & 
{\scriptsize$\pm$0.13} & {\scriptsize$\pm$0.59} \\

\multirow{2}{*}{Ann.} & 69.95 & 70.34 & 
70.28 & 69.95  \\
 & {\scriptsize$\pm$0.05} & {\scriptsize$\pm$0.14} & 
{\scriptsize$\pm$0.08} & {\scriptsize$\pm$0.07} \\

\hline

\end{tabular}
\end{center}
\label{table:mixcombT_GENE}
\end{table}

\textbf{Leveraging heterogeneous teachers.} We conducted experiments with heterogenous structure of teachers. As illustrated in Fig. \ref{table:mixcombT_GENE}, one better teacher does not guarantee a better student, which corroborates the previous studies \cite{cho2019efficacy}. Even though teachers have heterogeneous structures, they complement each other to improve the performance, which is shown with better performance than a model learned from scratch (Student).

\subsection{Effect of Mixup in KD}\label{Mixup_KD} 
In this section, we explore effects of mixup for learning from scratch and KD, which provides smoothness in training process. To analyze the interplay of mixup and KD, we utilize response based KD methods, including Base and Ann., which does not require to use additional weights and aids in more prominently showing the effects of interplay with mixup. Firstly, we train a model from scratch with mixup. Secondly, we train a student in KD with mixup. Also, to see the effects of smoothness by temperature in KD, we train students with different temperatures.

\begin{figure}[htb!]
\includegraphics[scale=0.43] {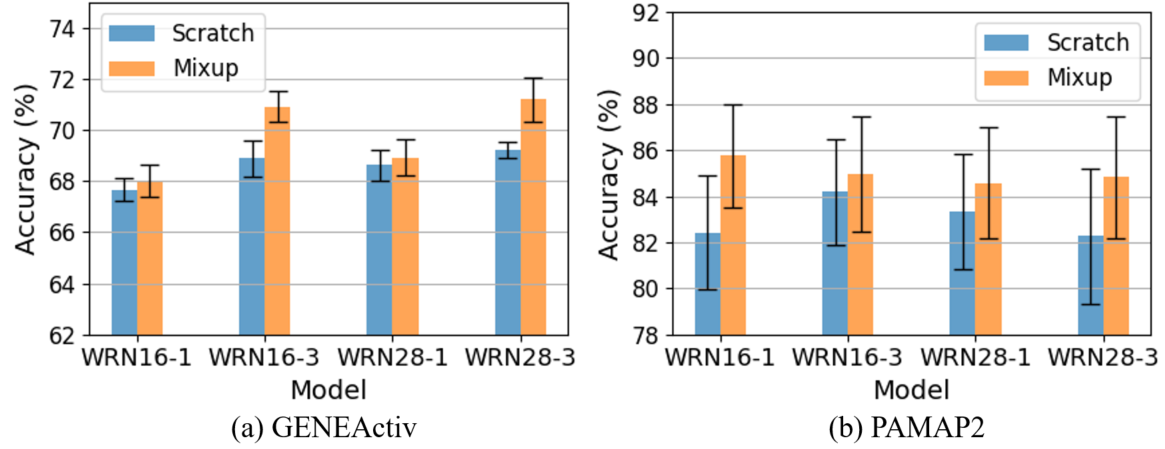} 
\centering
\caption{Results of various models trained from scratch with or without mixup.}
\label{figure:scratch_mixup}
\end{figure}

\begin{figure}[htb!]
\includegraphics[scale=0.54] {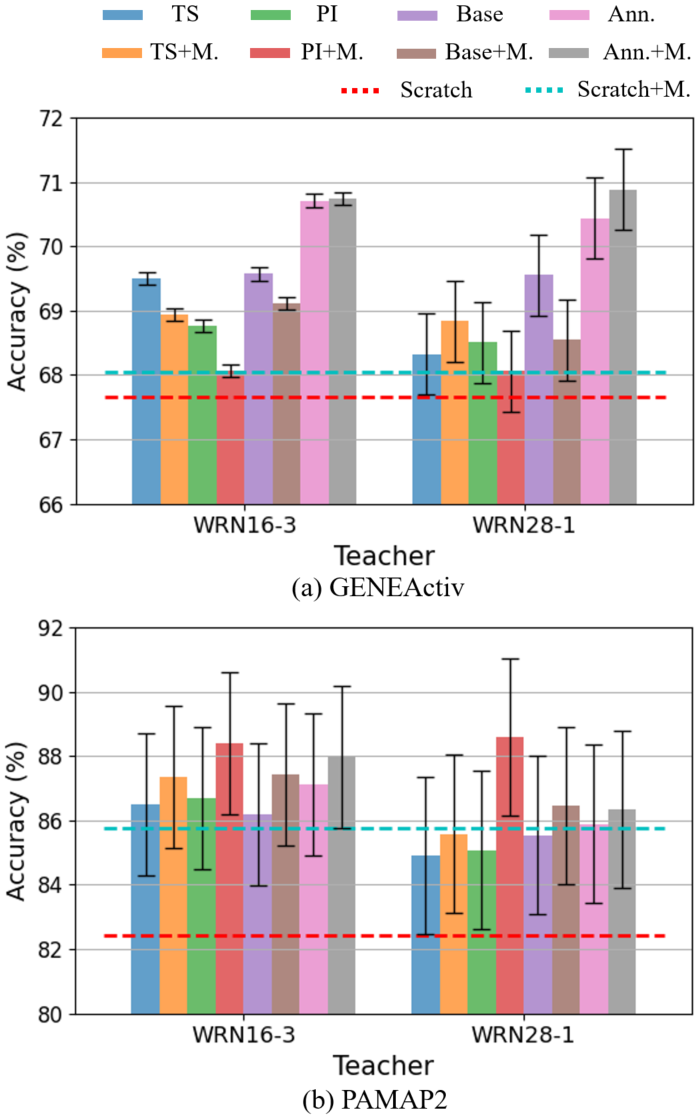} 
\centering
\caption{Results of various models trained with KD and mixup. TS and PI are results of students trained with KD. M. denotes using mixup.}
\label{figure:TmixupS}
\end{figure}

\begin{figure}[htb!]
\includegraphics[scale=0.43] {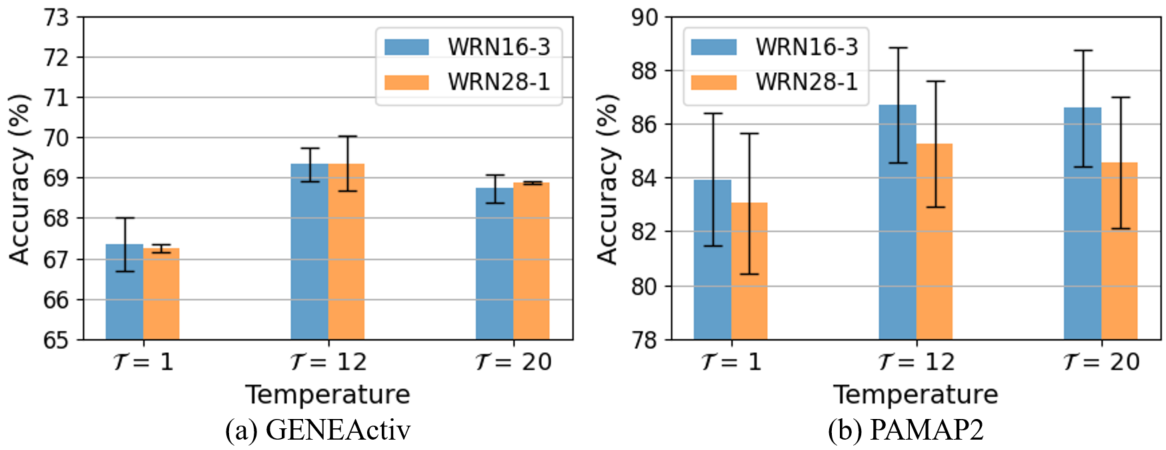} 
\centering
\caption{Results of various models with different temperature in KD.}
\label{figure:temperature}
\end{figure}

\begin{figure*}[htb!]
\includegraphics[scale=0.58] {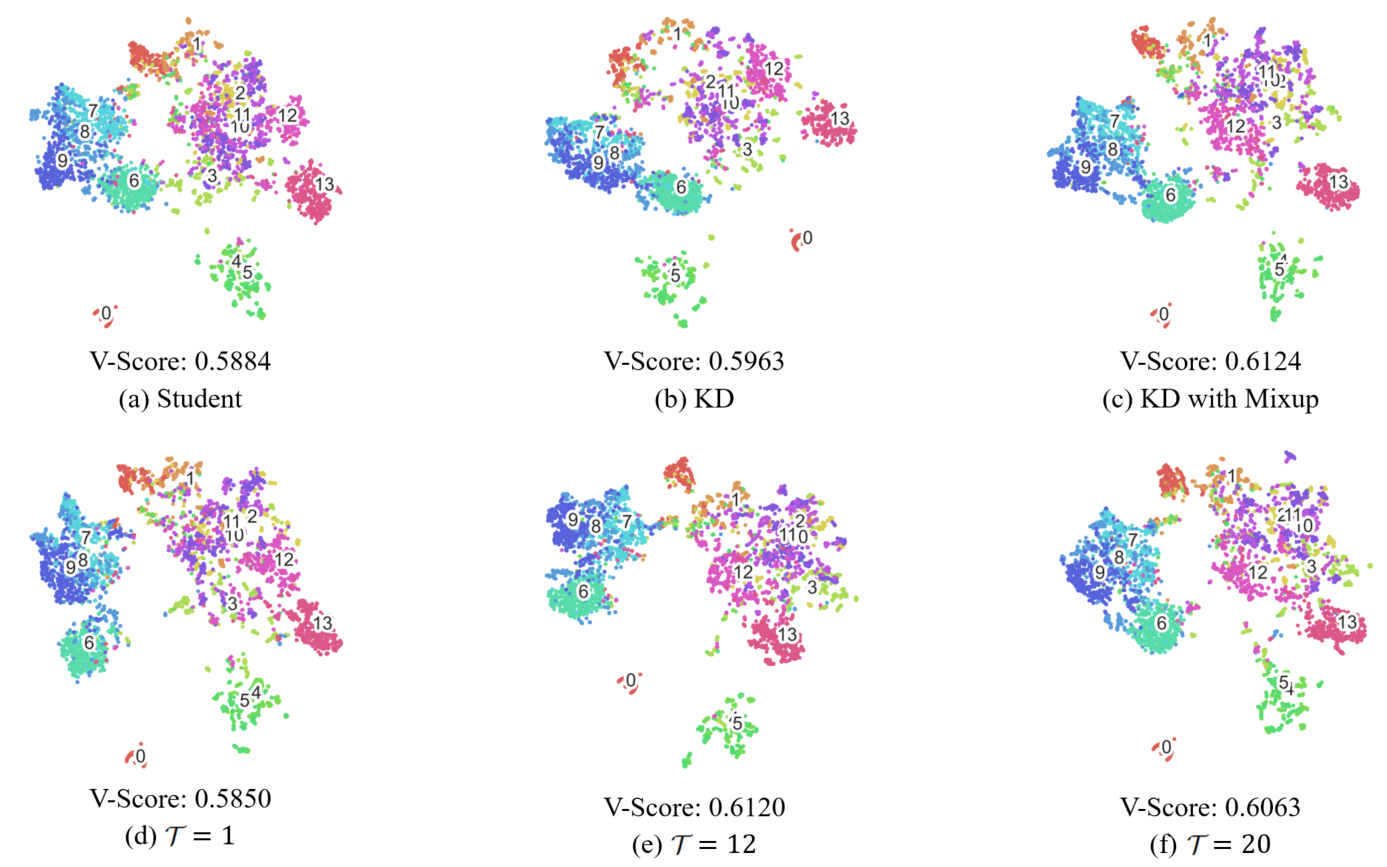} 
\centering
\caption{t-SNE plots of output for various models on GENEActiv. A teacher is WRN16-3 and a student is WRN16-1, which are trained with time-series data. ``Student'' is a model learned from scratch.}
\label{figure:tsne}
\end{figure*}

We trained various models from scratch with mixup, as illustrated in Fig. \ref{figure:scratch_mixup}. In all cases, models trained with mixup show better performance. In Fig. \ref{figure:TmixupS}, we show the results of various models trained with KD and mixup. WRN16-1 is used as a student. Mixup is applied to train a student in KD. In overall cases, with mixup generates better results. However, in some cases, the performance is worse than without mixup. This implies that mixup affects differently in KD compared to learning from scratch. Specifically, significant characteristics of input data, such as peaky points within a sample, can be softened because of blending different data for mixup, which was similar to results of injecting smoothness as addressed in previous study \cite{jeon2022kd}.
In all cases, Ann. shows better performance when mixup is added. This represents that topological features can complement time-series features to improve the performance. In details, persistence image representation can aid to preserve significant information, which generates synergetic effect with time-series features for classification.
We also trained models with different temperature hyper-parameters that can generate a smoothness effect for knowledge transfer. As shown in Fig. \ref{figure:temperature}, when $\mathcal{T}$=12, all cases show the best. Therefore, temperature can significantly affect to performance in KD.

We plot t-SNE with a WRN16-3 teacher and WRN16-1 student and measure the V-Score \cite{rosenberg2007v} of outputs from the penultimate layers in Fig. \ref{figure:tsne}. V-score is a metric to evaluate clustering, implying that a higher value is better clustering. 
For GENEActiv, classes from 0 to 5 are walking or running at different speeds. Class 7, 8, and 9 are activities related to hand motions such as brushing teeth and driving a car. Class 12 and 13 are walking up and down stairs, respectively.
When a student is trained with mixup, it generates a higher V-Score, compared to Student that is trained from scratch and results with conventional KD. 
Also, more distance between classes can be observed, which is measured with the V-Score and shown with the distance of the center point of the classes, particularly the gap between class 7, 8, and 9. In addition, some compacted points became more sparse, which is illustrated with class 12. For temperature, a high value of temperature provides more smoothness (soft knowledge) in KD, which can increase V-Score. When $\mathcal{T}$ is 12, the result shows the best, where the result is similar to the one of KD with mixup. When $\mathcal{T}$ is 1, the result is worse than learning from scratch. Thus, smoothness can affect the performance of KD at large. Based on these results, we can observe that injecting smoothness plays a key role in KD. That is, both mixup and temperature can significantly affect performance in distillation with generating soft knowledge, which can generate a synergistic effect to improve performance.

\textbf{Augmentations in KD.} Additionally, we conducted experiments with different augmentation methods (cutout \cite{devries2017improved} and cutmix \cite{yun2019cutmix}) in KD. 
The hyperparameter of cutout is 0.2. As explained in Table \ref{table:aug}, all augmentations show improved results for learning from scratch. However, with KD, mixup only achieves improvement while other augmentations show degradation. This corroborates the benefits of mixup in KD, explored in prior studies \cite{NEURIPS2022_57b53238, choi2023understanding, li2021smile, yang2022mixskd, xu2023computation, zhao2021data, zou2023benefits, beyer2022knowledge}.

\begin{table}[htb!]
\centering
\caption{Accuracy ($\%$) for different augmentations methods on GENEActiv. LS denotes learning from scratch.} 

\begin{center}
\begin{tabular}{c |c |c c c }

\hline
\centering

Method & Student & Mixup & Cutout & Cutmix \\ \hline
\multirow{2}{*}{LS} & \multirow{2}{*}{67.66{\scriptsize$\pm$0.45}} & 68.04{\scriptsize$\pm$0.63} & 68.67{\scriptsize$\pm$0.64} & 68.70{\scriptsize$\pm$0.94} \\
& & (\textcolor{ForestGreen}{0.38$\uparrow$}) & (\textcolor{ForestGreen}{1.01$\uparrow$}) & (\textcolor{ForestGreen}{1.04$\uparrow$})
\\
KD & \multirow{2}{*}{69.71{\scriptsize$\pm$0.38}} & 69.82{\scriptsize$\pm$0.24} & 65.79{\scriptsize$\pm$0.63} & 65.75{\scriptsize$\pm$0.65} \\
(WRN16-1)& & (\textcolor{ForestGreen}{0.11$\uparrow$}) & (\textcolor{Maroon}{3.92$\downarrow$}) & (\textcolor{Maroon}{3.96$\downarrow$})
\\
KD & \multirow{2}{*}{68.32{\scriptsize$\pm$0.63}} & 68.84{\scriptsize$\pm$0.23} & 65.03{\scriptsize$\pm$0.81} & 66.18{\scriptsize$\pm$0.44} \\
(WRN28-1)& & (\textcolor{ForestGreen}{0.52$\uparrow$}) & (\textcolor{Maroon}{3.29$\downarrow$}) & (\textcolor{Maroon}{2.14$\downarrow$})
\\
\hline

\end{tabular}
\end{center}
\label{table:aug}
\end{table}

\subsection{Teacher-Student with Mixup}

\begin{figure}[htb!]
\includegraphics[scale=0.52] {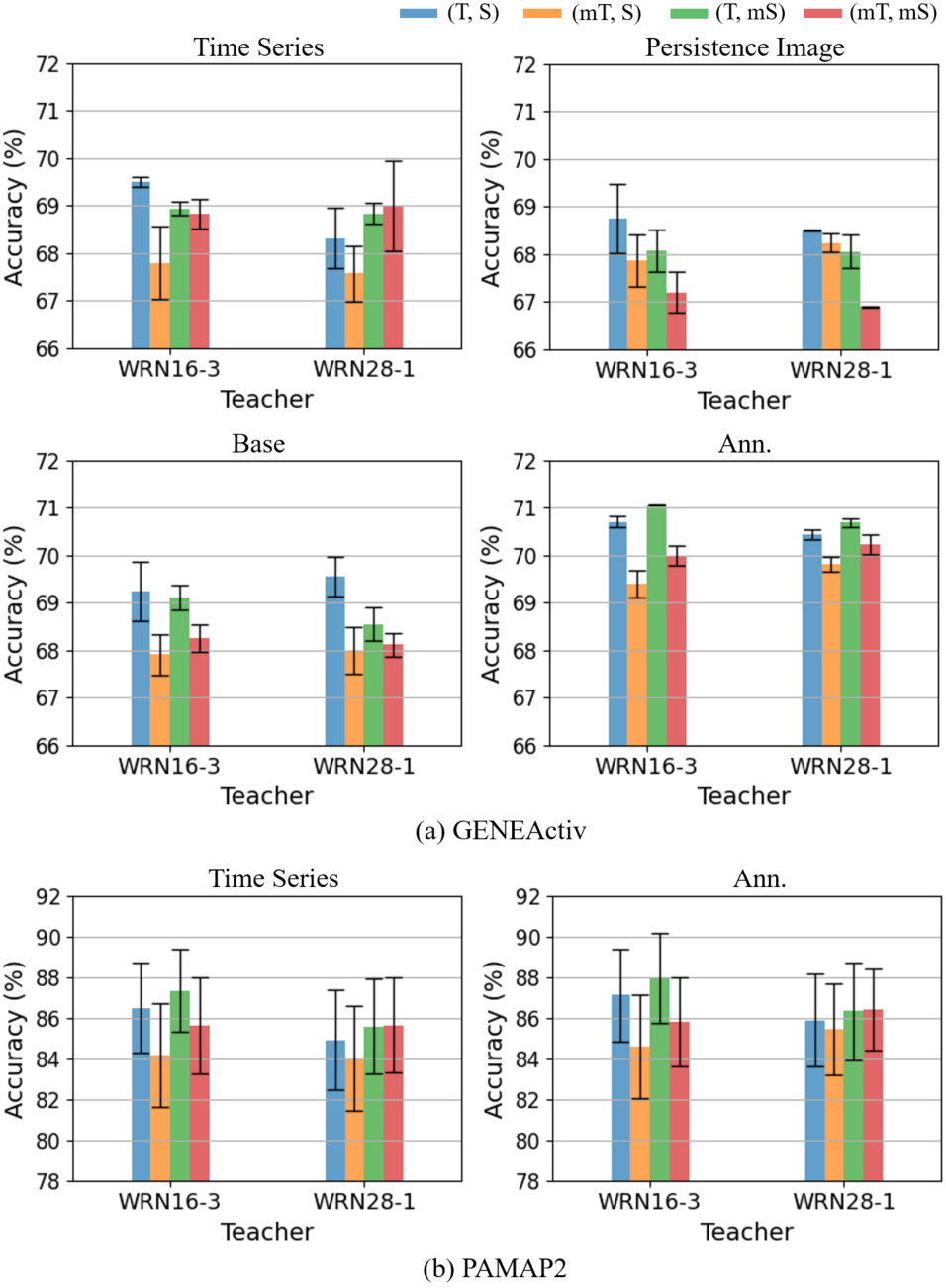} 
\centering
\caption{Results of various approaches in KD, trained with mixup. Brackets denote (Teacher, Student).}
\label{figure:mixupTS}
\end{figure}

To explore the effect of mixup-trained teachers as well as students, we set various combinations of using the augmentations in KD. Note, ``T'', ``S'', ``mT'' and ``mS'' denote a teacher model, a student model, a mixup-trained teacher model, and using mixup to train a student model. As explained in previous sections, WRN16-3 teachers generated a superior student compared to other combinations. On the other hand, WRN28-1 model learned from scratch showed less improvement with mixup than other capacity of models. For further analysis with mixup in KD, we use WRN16-3 and WRN28-1 for teachers and WRN16-1 for a student to consider different depth and width combinations of teacher-student networks and different effects on mixup in KD. As shown in Fig. \ref{figure:mixupTS}, Ann. shows the best among different approaches in KD. Students distilled by using PI alone and Base show worse performance than the one learned from scratch without using mixup. For Ann., when teachers are trained without mixup and a student is trained with mixup (T, mS), the student outperforms learning from scratch and other combinations of teacher-student trained with/without mixup.
These results represent that reducing knowledge gap with an annealing strategy (Ann.) is effective for applying mixup in KD to train a student with multiple teachers. Also, soft knowledge of topological persistence provided by mixup indeed aid to train a student.
In addition, this result corroborates the fact that the effects of mixup are similar to those of time domain augmentation methods, such as Gaussian noise, providing smoothness in KD, as analyzed in the previous study \cite{jeon2022kd}.

\subsection{Analysis of the Effects of Smoothness} 

\subsubsection{Analysis of Temperature with Mixup-trained Student}

\begin{figure}[htb!]
\includegraphics[scale=0.47] {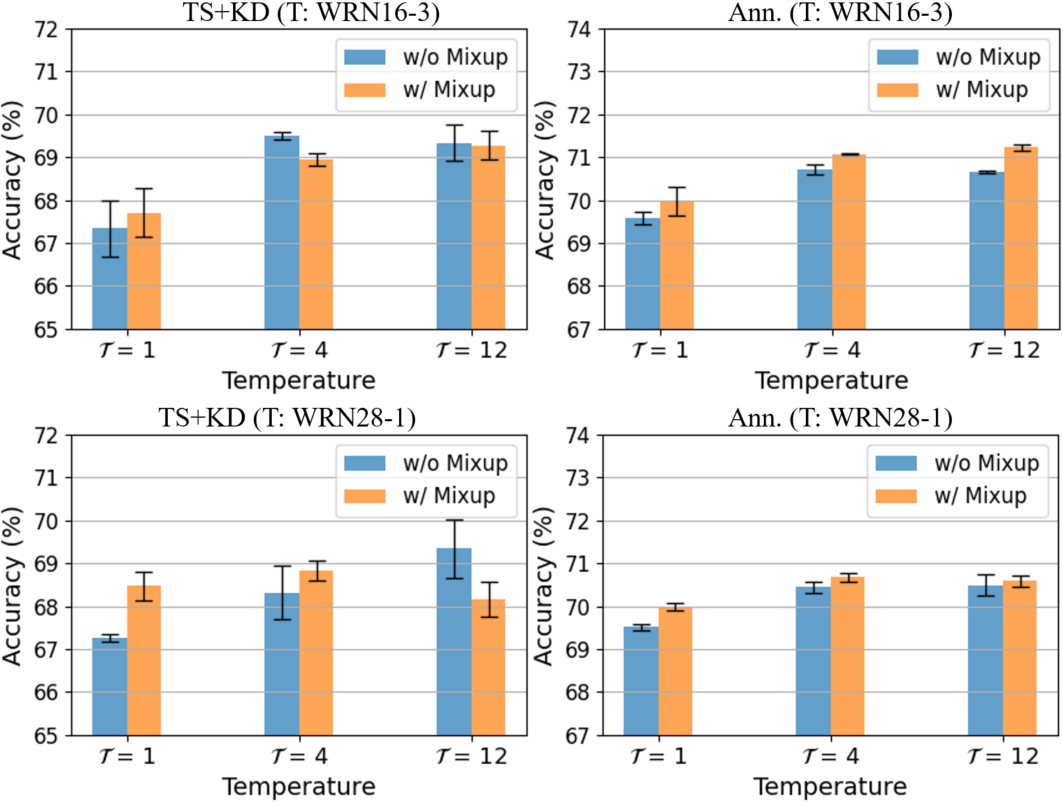} 
\centering
\caption{Results of various models with different temperature and mixup in KD on GENEActiv. Mixup is applied when a student is trained.}
\label{figure:temperature_mixup}
\end{figure}

\begin{figure}[htb!]
\includegraphics[scale=0.315] {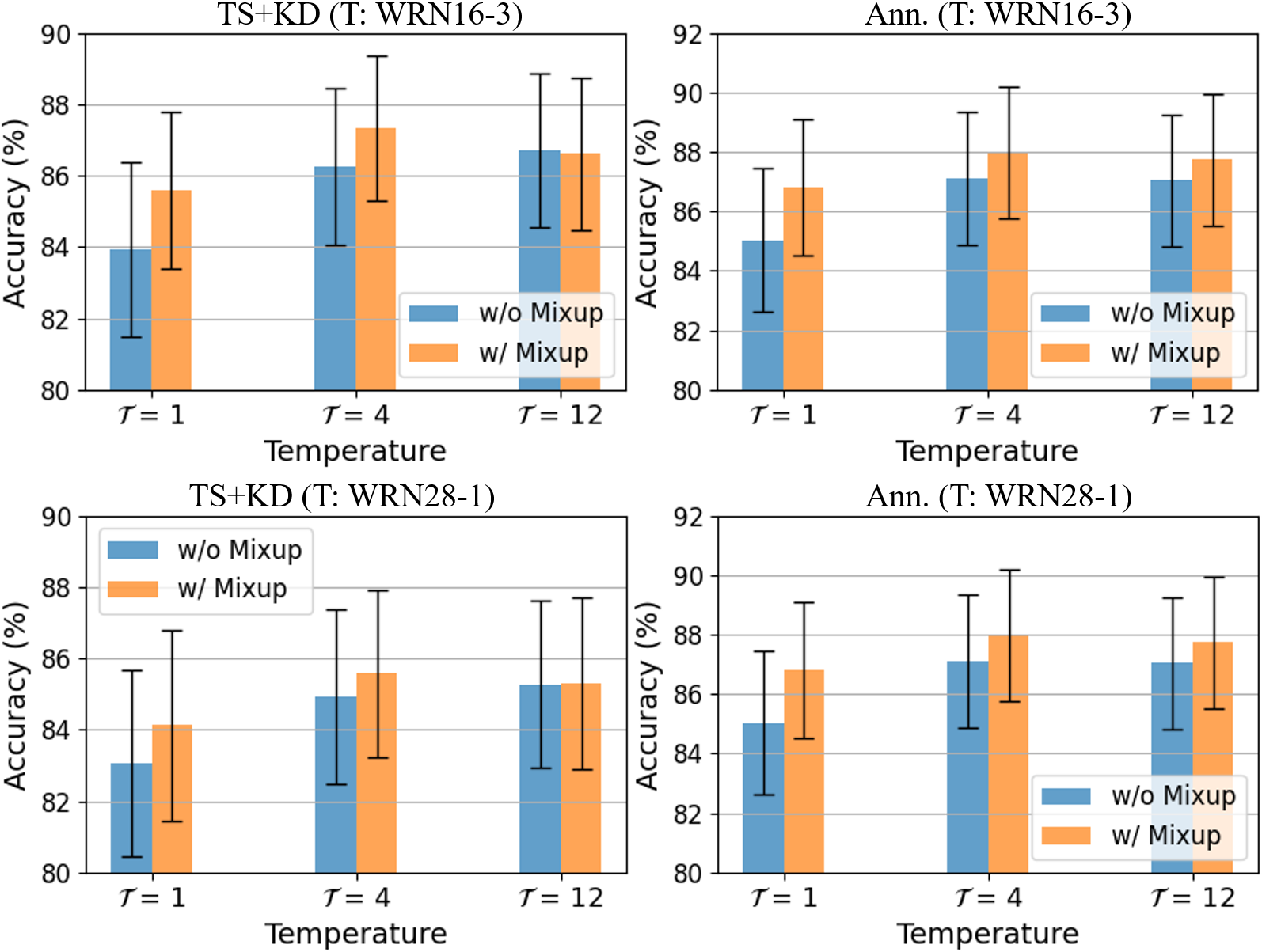} 
\centering
\caption{Results of various models with different temperature and mixup in KD on PAMAP2. Mixup is applied when a student is trained.}
\label{figure:temperature_mixup_p}
\end{figure}

In previous sections, we observed that both temperature and mixup inject smoothness into KD training process. To investigate the compatibility of smoothness with temperature and mixup, we evaluate KD with time-series data (TS+KD) and Ann. with different temperature parameters. The results of GENEActiv is illustrated in Fig. \ref{figure:temperature_mixup}. For TS+KD, when $\mathcal{T}$ is 1, with mixup improves the performance, implying that injecting smoothness can aid for training a student in KD. 
For both KD with time-series and Ann, in without mixup cases, it shows the best when $\mathcal{T}$ is 4 for WRN16-3 teacher and $\mathcal{T}$ is 12 for WRN16-3 teacher. With mixup, it shows the best when $\mathcal{T}$ is 12 for WRN16-3 teacher and $\mathcal{T}$ is 4 for WRN28-1 teacher, which are different from without mixup.
In Fig. \ref{figure:temperature_mixup_p}, for PAMAP2, KD with time-series data without mixup performs the best when $\mathcal{T}$ is 12. However, other results show their best when $\mathcal{T}$ is 4.
For both datasets, some accuracy results of KD with time-series and mixup are lower than those without mixup. This represents that excessive smoothness can hinder the training process in KD.
For Ann. with mixup outperforms without mixup in all cases. This implies that Ann. has better compatibility for utilizing mixup in KD and can allow more smoothness to improve performance than training with time-series alone.

\subsubsection{Partial Mixup}

To control the effects of smoothness on training procedures, we use PMU to alleviate excessive smoothness, which can degrade performance. We utilize different amounts of mixup pairs such as 0$\%$, 10$\%$, 50$\%$, and 100$\%$, where 0$\%$ means mixup is not applied and 100$\%$ denotes all samples of mixup pairs are used for training (FMU). Mixup is applied when a student is trained. As described in Table \ref{table:pmuGENE}, when teacher models are WRN16-3, less amounts of mixup pairs can distill a better student. When teacher models are WRN28-1, 50$\%$ of PMU shows the best. In Table \ref{table:pmuPAMAP2}, for PAMAP2, FMU shows the best. However, for WRN28-1, PMU with 10$\%$ of Ann. distills the best student. 
These results show that fewer mixup pairs can generate better performance. Also, if complexity of a dataset is high, mixup pairs contributes more to improving performance. On the other hand, KD with time-series data and Ann. have different optimal proportions of mixup pairs. This may be because Ann. uses both representations, including both time-series with 1D data and topological representations with 2D data, for training. Mixup influences different representations differently, so utilizing two teachers can provide more diverse relaxed knowledge for distillation, which is different from using one single teacher.


\begin{table}[htb!]
\centering
\caption{Accuracy ($\%$) with various mixup pair proportions on GENEActiv.}

\begin{center}
\begin{tabular}{c c |c c c c }

\hline
\centering


\multirow{2}{*}{Teachers} & \multirow{2}{*}{Method} & \multirow{2}{*}{No mixup} & \multirow{2}{*}{\shortstack{PMU \\ 0.1}} & \multirow{2}{*}{\shortstack{PMU \\ 0.5}} &  \multirow{2}{*}{FMU}\\
 & & & & & \\
\hline

\multirow{4}{*}{WRN16-3} & \multirow{2}{*}{TS+KD} &  \textbf{69.50} & 69.20 & 69.11 & 68.94 \\
& & {\scriptsize$\pm$0.10} & {\scriptsize$\pm$0.06} & {\scriptsize$\pm$0.27} &  {\scriptsize$\pm$0.15} \\
 & \multirow{2}{*}{Ann.} & 70.71 & \textbf{71.13} & 70.73 & 71.07 \\ 
 & & {\scriptsize$\pm$0.12} & {\scriptsize$\pm$0.14} & {\scriptsize$\pm$0.06} &  {\scriptsize$\pm$0.01} \\
 
 \hline
 
\multirow{4}{*}{WRN28-1} & \multirow{2}{*}{TS+KD} & 68.32 & 69.17 & \textbf{69.05} & 68.84 \\
& & {\scriptsize$\pm$0.63} & {\scriptsize$\pm$0.36} & {\scriptsize$\pm$0.15} &  {\scriptsize$\pm$0.23} \\
 & \multirow{2}{*}{Ann.} & 70.44 & 70.75 & \textbf{70.82} & 70.68 \\ 
 & & {\scriptsize$\pm$0.10} & {\scriptsize$\pm$0.02} & {\scriptsize$\pm$0.05} &  {\scriptsize$\pm$0.10} \\
 
\hline

\end{tabular}
\end{center}
\label{table:pmuGENE}
\end{table}


\begin{table}[htb!]
\centering
\caption{Accuracy ($\%$) with various mixup pair proportions on PAMAP2.}

\begin{center}
\begin{tabular}{c c |c c c c }

\hline
\centering


\multirow{2}{*}{Teachers} & \multirow{2}{*}{Method} & \multirow{2}{*}{No mixup} & \multirow{2}{*}{\shortstack{PMU \\ 0.1}} & \multirow{2}{*}{\shortstack{PMU \\ 0.5}} &  \multirow{2}{*}{FMU}\\
 & & & & & \\
\hline

\multirow{4}{*}{WRN16-3} & \multirow{2}{*}{TS+KD} & 86.50 & 86.75 & 86.05 & \textbf{87.34} \\
& & {\scriptsize$\pm$2.21} & {\scriptsize$\pm$2.10} & {\scriptsize$\pm$2.27} &  {\scriptsize$\pm$2.03} \\
 & \multirow{2}{*}{Ann.} & 87.12 & 87.63 & 87.54 & \textbf{87.98} \\ 
 & & {\scriptsize$\pm$2.26} & {\scriptsize$\pm$2.35} & {\scriptsize$\pm$2.34} &  {\scriptsize$\pm$2.21} \\
 
 \hline
 
\multirow{4}{*}{WRN28-1} & \multirow{2}{*}{TS+KD} & 84.92 & 85.42 & 85.36 & \textbf{85.58} \\
& & {\scriptsize$\pm$2.45} & {\scriptsize$\pm$2.30} & {\scriptsize$\pm$2.48} &  {\scriptsize$\pm$2.26} \\
 & \multirow{2}{*}{Ann.} & 85.89 & \textbf{86.69} & 86.47 & 86.35 \\ 
 & & {\scriptsize$\pm$2.26} & {\scriptsize$\pm$2.20} & {\scriptsize$\pm$2.29} &  {\scriptsize$\pm$2.39} \\
 
\hline

\end{tabular}
\end{center}
\label{table:pmuPAMAP2}
\end{table}

\subsection{Mixup for Different Teachers}

Since two teachers can provide different effects on distillation, we use different hyper-parameters for mixup to knowledge transfer from two teachers when a student is trained in KD. We utilize Ann. that shows the best in most of the cases presented in the previous sections. Note, $\alpha_1$ and $\alpha_2$ are hyper-parameters of mixup for Teacher1 and Teacher2. As summarized in Table \ref{table:alphaGENE} and \ref{table:alphaPAMAP2}, applying different mixup hyper-parameters can distill a better student.


\begin{table}[htb!]
\centering
\caption{Accuracy ($\%$) with various hyper-parameter pairs of mixup for teachers on GENEActiv. Ann. is used for KD.}

\begin{center}
\begin{tabular}{c c |c c }

\hline
\centering

\multirow{2}{*}{$\alpha_1$} & \multirow{2}{*}{$\alpha_2$} & \multicolumn{2}{c}{Teachers} \\

& & WRN16-3 & WRN28-1 \\ \hline
0.1 & 0.1 & 70.72{\scriptsize$\pm$0.06} & 70.88{\scriptsize$\pm$0.04} \\
0.1 & 0.15 & 70.93{\scriptsize$\pm$0.11} & 70.79{\scriptsize$\pm$0.12} \\
0.15 & 0.1 & 70.99{\scriptsize$\pm$0.03} & 70.88{\scriptsize$\pm$0.18} \\
0.15 & 0.15 & 70.96{\scriptsize$\pm$0.16} & \textbf{71.16}{\scriptsize$\pm$0.05} \\
0.15 & 0.2 & 71.07{\scriptsize$\pm$0.14} & 71.01{\scriptsize$\pm$0.16} \\
0.2 & 0.15 & \textbf{71.22}{\scriptsize$\pm$0.12} & 71.00{\scriptsize$\pm$0.07} \\
0.2 & 0.2 & 71.17{\scriptsize$\pm$0.22} & 70.93{\scriptsize$\pm$0.21} \\

\hline

\end{tabular}
\end{center}
\label{table:alphaGENE}
\end{table}


\begin{table}[htb!]
\centering
\caption{Accuracy ($\%$) with various hyper-parameter pairs of mixup for teachers on PAMAP2. Ann. is used for KD.}

\begin{center}
\begin{tabular}{c c |c c }

\hline
\centering

\multirow{2}{*}{$\alpha_1$} & \multirow{2}{*}{$\alpha_2$} & \multicolumn{2}{c}{Teachers} \\

& & WRN16-3 & WRN28-1 \\ \hline
0.1 & 0.1 & 87.98{\scriptsize$\pm$2.21} & 86.35{\scriptsize$\pm$2.39} \\
0.1 & 0.15 & \textbf{87.99}{\scriptsize$\pm$2.29} & \textbf{86.72}{\scriptsize$\pm$2.41} \\
0.15 & 0.1 & 87.94{\scriptsize$\pm$2.26} & 86.00{\scriptsize$\pm$2.43} \\
0.15 & 0.15 & 87.67{\scriptsize$\pm$2.21} & 86.70{\scriptsize$\pm$2.35} \\

\hline

\end{tabular}
\end{center}
\label{table:alphaPAMAP2}
\end{table}


As depicted in Table \ref{table:diffmixupGENE} and \ref{table:diffmixupPAMAP2}, we evaluate with different teachers having different architectural designs of depth and width for networks. Mix. denotes applying mixup for training a student. $\alpha$ of mixup is 0.1. When $\alpha$ is applied differently for teachers (diff. $\alpha$), ($\alpha_1$, $\alpha_2$) is (0.15, 0.2) for GENEActiv and (0.1, 0.15) for PAMAP2. In all cases, applying different mixup hyper-parameters can distill a better student.


\begin{table}[htb!]
\centering
\renewcommand{\tabcolsep}{1.0mm} 
\caption{Accuracy ($\%$) with various knowledge distillation methods and different hyper-parameter of mixup for teachers on GENEActiv.}

\begin{center}
 \scalebox{0.92}{
\begin{tabular}{c c c|c |c |c |c }

\hline
\centering

\multirow{3}{*}{\shortstack{Teacher1 \\ (1D CNNs)}}& \multirow{3}{*}{\shortstack{Teacher2 \\ (2D CNNs)}} & \multirow{3}{*}{\shortstack{Student \\ (1D CNNs)}} &  \multicolumn{4}{c}{TS+PI} \\
\cline{4-7}
 & & & \multirow{2}{*}{Base} & \multirow{2}{*}{Ann.} & Ann. & Ann. \\
 & & & & & +Mix. & +Mix. (diff. $\alpha$) \\
\hline

WRN16-1 & WRN28-1 &  \multirow{9}{*}{\shortstack{WRN16-1 \\ (0.06M \\ 67.66)}} & \multirow{3}{*}{\shortstack{68.71 \\ {\scriptsize$\pm$0.36}}} & \multirow{3}{*}{\shortstack{69.95 \\ {\scriptsize$\pm$0.05}}} & \multirow{3}{*}{\shortstack{70.67 \\ {\scriptsize$\pm$0.05}}} & \multirow{3}{*}{\shortstack{\textbf{70.92} \\ {\scriptsize$\pm$0.24}}} \\
(0.06M, & (0.4M, &   & & & & \\
67.66) & 59.45) &  & & & &  \\

WRN28-1 & WRN28-3 &  & \multirow{3}{*}{\shortstack{68.26 \\ {\scriptsize$\pm$0.13}}} & \multirow{3}{*}{\shortstack{70.28 \\ {\scriptsize$\pm$0.08}}} & \multirow{3}{*}{\shortstack{70.74 \\ {\scriptsize$\pm$0.15}}} & \multirow{3}{*}{\shortstack{\textbf{70.86} \\ {\scriptsize$\pm$0.13}}} \\
(0.1M, & (3.3M, &  & & & & \\
68.63) & 59.69) & & & & &  \\

WRN40-1 & WRN28-3 &  & \multirow{3}{*}{\shortstack{68.90 \\ {\scriptsize$\pm$0.50}}} & \multirow{3}{*}{\shortstack{70.49 \\ {\scriptsize$\pm$0.05}}} & \multirow{3}{*}{\shortstack{70.91 \\ {\scriptsize$\pm$0.05}}} & \multirow{3}{*}{\shortstack{\textbf{71.21} \\ {\scriptsize$\pm$0.06}}} \\
(0.2M, & (3.3M, &  & & & & \\
69.05) & 59.69) & & & & &  \\

\hline

\end{tabular}
}
\end{center}
\label{table:diffmixupGENE}
\end{table}



\begin{table}[htb!]
\centering
\renewcommand{\tabcolsep}{1.0mm} 
\caption{Accuracy ($\%$) with various knowledge distillation methods and different hyper-parameter of mixup for teachers on PAMAP2.}

\begin{center}
 \scalebox{0.92}{
\begin{tabular}{c c c|c |c |c |c }

\hline
\centering

\multirow{3}{*}{\shortstack{Teacher1 \\ (1D CNNs)}}& \multirow{3}{*}{\shortstack{Teacher2 \\ (2D CNNs)}} & \multirow{3}{*}{\shortstack{Student \\ (1D CNNs)}} &  \multicolumn{4}{c}{TS+PI} \\
\cline{4-7}
 & & & \multirow{2}{*}{Base} & \multirow{2}{*}{Ann.} & Ann. & Ann. \\
 & & & & & +Mix. & +Mix. (diff. $\alpha$) \\
\hline

WRN16-1 & WRN28-1 &  \multirow{9}{*}{\shortstack{WRN16-1 \\ (0.06M \\ 82.99)}} & \multirow{3}{*}{\shortstack{85.78 \\ {\scriptsize$\pm$2.29}}} & \multirow{3}{*}{\shortstack{85.33 \\ {\scriptsize$\pm$2.22}}} & \multirow{3}{*}{\shortstack{86.47 \\ {\scriptsize$\pm$2.35}}} & \multirow{3}{*}{\shortstack{\textbf{87.09} \\ {\scriptsize$\pm$2.16}}} \\
(0.06M, & (0.4M, &   & & & & \\
85.27) & 87.45) &  & & & &  \\

WRN28-3 & WRN28-1 &  & \multirow{3}{*}{\shortstack{85.69 \\ {\scriptsize$\pm$2.41}}} & \multirow{3}{*}{\shortstack{85.59 \\ {\scriptsize$\pm$2.28}}} & \multirow{3}{*}{\shortstack{87.06 \\ {\scriptsize$\pm$2.17}}} & \multirow{3}{*}{\shortstack{\textbf{87.80} \\ {\scriptsize$\pm$2.09}}} \\
(1.1M, & (0.4M, &  & & & & \\
84.46) & 87.45) & & & & &  \\

WRN16-3 & WRN28-1 &  & \multirow{3}{*}{\shortstack{85.48 \\ {\scriptsize$\pm$2.37}}} & \multirow{3}{*}{\shortstack{85.82 \\ {\scriptsize$\pm$2.26}}} & \multirow{3}{*}{\shortstack{86.80 \\ {\scriptsize$\pm$2.23}}} & \multirow{3}{*}{\shortstack{\textbf{87.29} \\ {\scriptsize$\pm$2.20}}} \\
(0.5M, & (0.4M, &  & & & & \\
85.80) & 87.45) & & & & &  \\

\hline

\end{tabular}
}
\end{center}
\label{table:diffmixupPAMAP2}
\end{table}


To figure out if using different mixup hyper-parameters can complement the partial mixup method, we apply different proportions of mixup pairs for training a student with different mixup hyper-parameters. In Table \ref{table:alphapmGENE}, FMU shows the best for both cases of teachers. With small proportions of mixup pairs, a large degradation of performance is shown, where the results are lower than training without mixup. When the complexity of the dataset is low and the size of the model is small, partial mixup can yield an adverse effect on training a student, which may produce pairs of inputs that are not expressive enough to learn. In Table \ref{table:alphapmPAMAP2}, 50$\%$ of mixup pairs show the best. These results imply that using the proper mixup pair proportion for training a student is important to improve their performance in KD. Also, considering the effects on different relaxed knowledge of a mixup from two teachers can generate a better student. 



\begin{table}[htb!]
\centering
\caption{Accuracy ($\%$) with various hyper-parameter pairs of mixup on GENEActiv. Ann. is used for KD.}

\begin{center}
\begin{tabular}{c c c |c c }

\hline
\centering

\multirow{2}{*}{$\alpha_1$} & \multirow{2}{*}{$\alpha_2$} & \multirow{2}{*}{Mixup} & \multicolumn{2}{c}{Teachers} \\

& & & WRN16-3 & WRN28-1 \\ \hline
0.15 & 0.2 & FMU & 71.07{\scriptsize$\pm$0.14} & \textbf{71.01}{\scriptsize$\pm$0.16} \\
0.15 & 0.2 & PMU(50$\%$) & 70.57{\scriptsize$\pm$0.17} & 70.46{\scriptsize$\pm$0.10} \\
0.15 & 0.2 & PMU(10$\%$) & 70.55{\scriptsize$\pm$0.14} & 70.73{\scriptsize$\pm$0.24} \\
0.2 & 0.15 & FMU & \textbf{71.22}{\scriptsize$\pm$0.12} & 71.00{\scriptsize$\pm$0.07} \\
0.2 & 0.15 & PMU(50$\%$) & 70.37{\scriptsize$\pm$0.05} & 70.42{\scriptsize$\pm$0.03} \\
0.2 & 0.15 & PMU(10$\%$) & 70.64{\scriptsize$\pm$0.04} & 70.38{\scriptsize$\pm$0.23} \\

\hline

\end{tabular}
\end{center}
\label{table:alphapmGENE}
\end{table}


\begin{table}[htb!]
\centering
\caption{Accuracy ($\%$) with various hyper-parameter pairs of mixup on PAMAP2. Ann. is used for KD.}

\begin{center}
\begin{tabular}{c c c |c c }

\hline
\centering

\multirow{2}{*}{$\alpha_1$} & \multirow{2}{*}{$\alpha_2$} & \multirow{2}{*}{Mixup} & \multicolumn{2}{c}{Teachers} \\

& & & WRN16-3 & WRN28-1 \\ \hline
0.1 & 0.15 & FMU & 87.99{\scriptsize$\pm$2.29} & 86.72{\scriptsize$\pm$2.41} \\
0.1 & 0.15 & PMU(50$\%$) & \textbf{88.13}{\scriptsize$\pm$2.19} & \textbf{86.73}{\scriptsize$\pm$2.23} \\
0.1 & 0.15 & PMU(10$\%$) & 87.88{\scriptsize$\pm$2.29} & 86.68{\scriptsize$\pm$2.26} \\

\hline

\end{tabular}
\end{center}
\label{table:alphapmPAMAP2}
\end{table}


\subsection{Analysis of Optimized Solution}

\subsubsection{Parametric Plots}

A solution space comparison for two models can give a valuable understanding of their behavior in training or testing and how these models are related. One of the useful tools for the analysis is the parametric plot that has been widely studied \cite{goodfellow2014qualitatively, zhu2022rethinking, keskar2017on}.

\begin{figure}[htb!]
\includegraphics[scale=0.405] {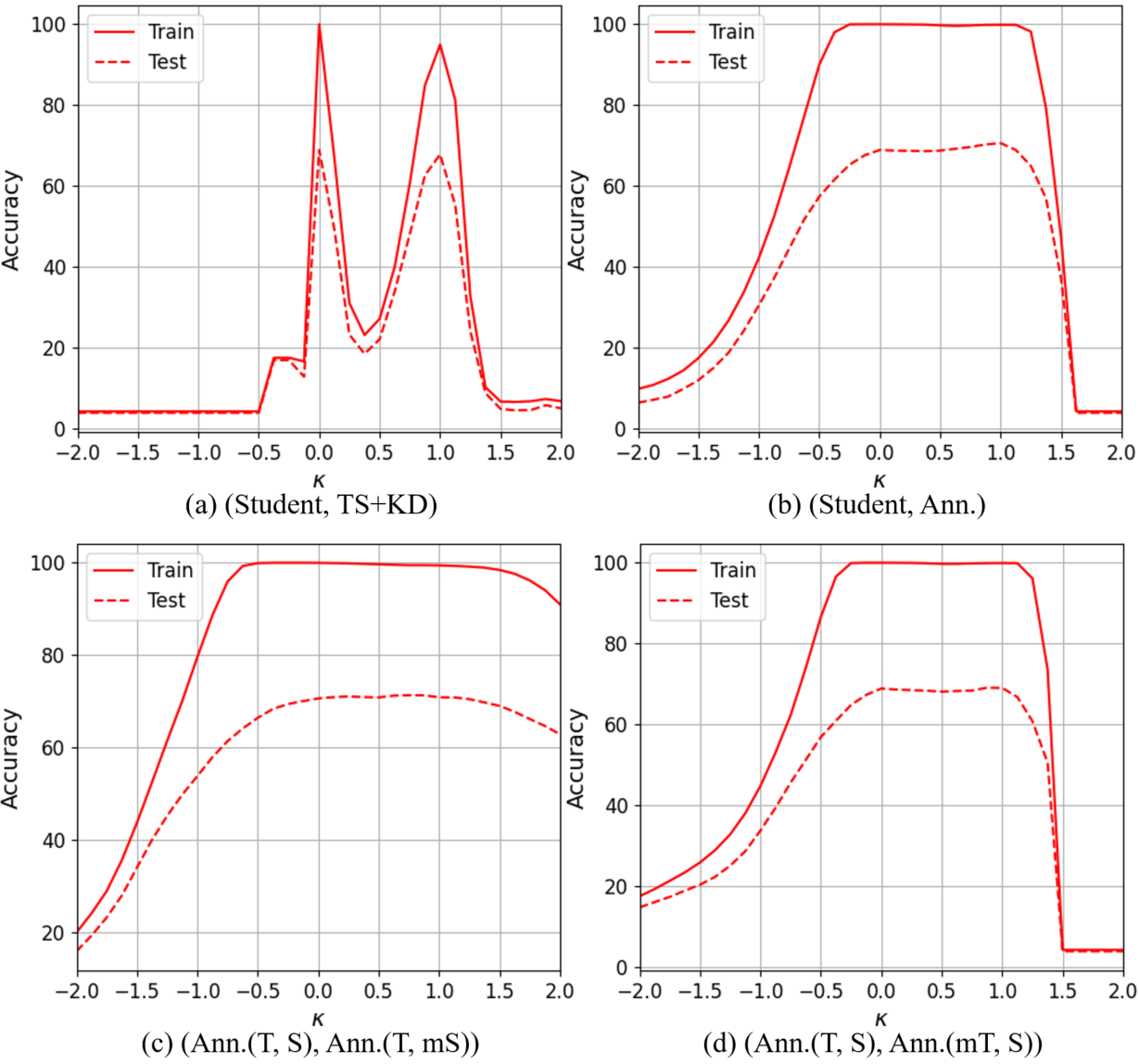} 
\centering
\caption{Parametric plots with accuracy ($\%$) for various pairs of models on GENEActiv. Brackets denote solutions ($z^{\ast}_a$, $z^{\ast}_b$). $\kappa$ = 0 implies to $z^{\ast}_a$ and $\kappa$ = 1 to $z^{\ast}_b$. ``Student'' is a model learned from scratch.}
\label{figure:param_plot}
\end{figure}

In Fig. \ref{figure:param_plot}, we plot classification accuracy for with function $\psi((1-\kappa) z^{\ast}_a + \kappa z^{\ast}_b)$ for $\kappa \in$ [-2, 2], where $z^{\ast}_a$ and $z^{\ast}_b$ are different solutions. Teachers are WRN16-3 and students are WRN16-1, which produced the best overall performance in the previous sections.
In Fig. \ref{figure:param_plot}(a), when $\kappa$ is 0.5, the accuracy of training and testing is lower than 30$\%$, which represents that the solution spaces of learning from scratch and KD with time-series data are different, whereas the result in Fig. \ref{figure:param_plot}(b) shows approximately 70$\%$ at $\kappa$ = 0.5. This implies that the solution space of Ann. is similar to that of Student. As illustrated in Fig. \ref{figure:param_plot}(c), it shows more flattened results. The result at around $\kappa$ = 1.0 shows a more gentle slope than the one at $\kappa$ = 0, which indicates that using mixup to train a student in KD leads to get benefits for failure prediction and mitigates reliable over-fitting.  When a mixup trained teacher is used, the student's solution space is similar to that of a non-mixup trained teacher. Based on (c) and (d), we can observe that utilizing mixup trained students (T, mS) leads to a better solution space that is relatively less susceptible to perturbations than using mixup trained teachers (mT, S).

\subsubsection{Mixup Hyper-parameter $\alpha$} To explore the performance on $\alpha$ of mixup and its sensitivity, we train various models with learning from scratch, KD, and Ann. using different settings of $\alpha$, which is described in Table \ref{table:valphaGENE}. 
The optimal $\alpha$ parameters for models trained with time-series and topological persistence are different. When $\alpha$ value is between the optimal one of TS and PI ($\alpha \in $ [0.1, 0.4]), Ann. performs better than training with the other value ($\alpha$ = 0.05). Therefore, setting the proper $\alpha$ leads to getting the best performance, and an intermediate $\alpha$ can generate the best performance when different teachers are applied.

\begin{table}[htb!]
\begin{center}
\caption{Test accuracy ($\%$) under different settings of $\alpha$ on GENEActiv. WRN16-1 is used for learning from scratch and a student.}
\label{table:valphaGENE}
\begin{tabular}{c|c|cccc}
\hline
\multicolumn{2}{c|}{\multirow{2}{*}{Method}} & \multicolumn{4}{c}{Mixup $\alpha$} \\  \cline{3-6}
\multicolumn{2}{c|}{} & 0.05 & 0.1 & 0.2 & 0.4 \\ \hline 
\multirow{4}{*}{\rotatebox[origin=c]{90}{
Scratch}} & \multirow{2}{*}{TS} & 67.99 & 68.04 & 69.28 & \textbf{69.35} \\
 & & {\scriptsize$\pm$0.41} & {\scriptsize$\pm$0.63} & {\scriptsize$\pm$0.19} & {\scriptsize$\pm$0.52}  \\
& \multirow{2}{*}{PI} &  59.23 & 59.08 & \textbf{59.71} & 59.47 \\ 
 & & {\scriptsize$\pm$0.41} & {\scriptsize$\pm$0.77} & {\scriptsize$\pm$0.58} & {\scriptsize$\pm$0.19}  \\ \hline
\multirow{6}{*}{\rotatebox[origin=c]{90}{KD (16-3)}} & \multirow{2}{*}{TS} & 69.02 & 68.94 & 69.15 & \textbf{69.39} \\
 & & {\scriptsize$\pm$0.22} & {\scriptsize$\pm$0.15} & {\scriptsize$\pm$0.13} & {\scriptsize$\pm$0.21}  \\
& \multirow{2}{*}{PI} & 67.31 & \textbf{68.08} & 66.77 & 68.02 \\
 & & {\scriptsize$\pm$0.28} & {\scriptsize$\pm$0.44} & {\scriptsize$\pm$0.66} & {\scriptsize$\pm$0.35}  \\
 & \multirow{2}{*}{Ann.} & 70.63 & 70.72 & 71.17 & \textbf{71.35} \\
 & & {\scriptsize$\pm$0.03} & {\scriptsize$\pm$0.06} & {\scriptsize$\pm$0.22} & {\scriptsize$\pm$0.14}  \\ \hline
 \multirow{6}{*}{\rotatebox[origin=c]{90}{KD (28-1)}} & \multirow{2}{*}{TS} & 68.95 & 68.84 & 68.74 & \textbf{69.16} \\
 & & {\scriptsize$\pm$0.44} & {\scriptsize$\pm$0.23} & {\scriptsize$\pm$0.39} & {\scriptsize$\pm$0.55}  \\
& \multirow{2}{*}{PI} & 67.77 & \textbf{68.06} & 67.92 & 67.83 \\
 & & {\scriptsize$\pm$0.50} & {\scriptsize$\pm$0.34} & {\scriptsize$\pm$0.49} & {\scriptsize$\pm$0.28}  \\ 
 & \multirow{2}{*}{Ann.} & 70.81 & 70.88 & \textbf{70.93} & 70.76 \\
 & & {\scriptsize$\pm$0.26} & {\scriptsize$\pm$0.04} & {\scriptsize$\pm$0.21} & {\scriptsize$\pm$0.19}  \\ \hline
\end{tabular}
\end{center}
\end{table}

\section{Discussion} \label{sec:discussion}
We explored the interplay between mixup and KD on diverse strategies with multimodal representations including topological features for wearable sensor data analysis. To achieve more improved synergistic effects, partial mixup can be utilized, which prevents excessive smoothing effects that generate degradation. 
As an extended research, these strategies introduced in this paper are applicable to diverse computer vision tasks \cite{han2024asymmetric, dong2022adaptive}, such as image recognition, object tracking and detection, and segmentation. For example, when a model for image recognition is trained with our strategy, the trained model can be utilized as a backbone model in a framework for many different computer vision tasks. Also, this study can be explored on vision based or different types of sensor signal, using motion capture or ECG, based human activity recognition. These can be more investigated as a future work.


\section{Conclusion} \label{sec:conclu}
In this paper, we explored the role of mixup in topological based KD with different approaches. We confirmed that mixup and temperature in KD have a connecting link that imposes smoothness for training process. Excessive smoothness produced inferior supervision that hinders training a student in KD. We observed that utilizing topological features can complement time-series to improve the end performance. Also, using topological persistence showed better compatibility when using mixup in KD.

Further, two teachers transfer different statistical knowledge so that their optimal parameters for augmentation in distillation can be different, where teachers are trained with time-series and topological features, respectively. We would like to extend a framework using multiple teachers to find optimal hyper-parameters of mixup and partial mixup adaptively, considering different statistical characteristics of teachers. In addition, our findings provide insights for developing further advanced distillation methods for various fields including wearable sensor data analysis and computer vision tasks.

\section*{Acknowledgment}
This work was supported in part by the National Institutes of Health under Grant R01GM135927 as part of the Joint DMS/NIGMS Initiative to Support Research at the Interface of the Biological and Mathematical Sciences, NSF grant 2200161, and in part by Seoul National University of Science and Technology.









\bibliographystyle{elsarticle-num} 
\bibliography{reference}

\end{document}